%% file: main.tex
\documentclass[letterpaper]{article} 
\usepackage{aaai23}
\usepackage{times}  
\usepackage{helvet}  
\usepackage{courier}  
\usepackage[hyphens]{url}  
\usepackage{graphicx} 
\usepackage{natbib}  
\usepackage{caption} 
\usepackage{algorithmic}
\usepackage{amsmath}
\usepackage{multirow}
\usepackage{threeparttable}
\usepackage{subfig}
\usepackage{makecell}
\usepackage{enumitem}
\usepackage{newfloat}
\usepackage{listings}
\usepackage{lineno}
\linenumbers

\DeclareCaptionStyle{ruled}{labelfont=normalfont,labelsep=colon,strut=off} 
\usepackage[accsupp]{axessibility}  
\usepackage{xcolor}
\usepackage{color, colortbl}

\usepackage{cite}
\usepackage{pifont}%

\usepackage{makecell}

\usepackage[utf8]{inputenc}        
\usepackage[T1]{fontenc}    
\usepackage{hyperref}        
\usepackage{url}            
\usepackage{booktabs}       
\usepackage{amsfonts}       
\usepackage{nicefrac}       
\usepackage{microtype}      
\usepackage{cleveref}
\usepackage{pdfpages}
\usepackage{float}
\usepackage{hhline}
\usepackage{array}
\usepackage{varwidth}
\usepackage{multirow}
\usepackage{makecell}
\usepackage{tabularx}
\usepackage{amsmath}
\usepackage[linesnumbered,ruled]{algorithm2e}
\usepackage{subfig}
\usepackage{graphicx}

\definecolor{mygreen}{RGB}{0 139 69}
\definecolor{mygreen2}{RGB}{0 205 0}
\definecolor{myred}{RGB}{205 38 38}
\definecolor{TartOrange}{HTML}{ff2e35}
\definecolor{Orange}{HTML}{ff7825}
\definecolor{Mango}{HTML}{ffc013}
\definecolor{AppleGreen}{HTML}{7cb81b}
\definecolor{Blue}{HTML}{1173b0}
\definecolor{BdazzledBlue}{HTML}{2e58a5}
\definecolor{Purple}{HTML}{5b3590}
\definecolor{Sunglow}{HTML}{FFCA3A}
\definecolor{Gray}{gray}{0.9}
\hypersetup{
	colorlinks=true,
	urlcolor=violet,
	citecolor=blue,
}

\newcommand{\cifar}{\textsc{Cifar-10}\xspace}
\newcommand{\imagenet}{\textsc{ImageNet}\xspace}
\newcommand{\cifarh}{\textsc{Cifar-100}\xspace}

\newcommand{\beyond}{\textsc{Beyond}\xspace}

\title{Be Your Own Neighborhood: Detecting Adversarial Example by the Neighborhood Relations Built on Self-Supervised Learning}

\author{
    Zhiyuan He \textsuperscript{\rm 1}\equalcontrib,
    Yijun Yang \textsuperscript{\rm 1}\equalcontrib,
    Pin-Yu Chen \textsuperscript{\rm 2},
    Qiang Xu \textsuperscript{\rm 1},
    Tsung-Yi Ho \textsuperscript{\rm 1}
}
\affiliations{
    \textsuperscript{\rm 1}Department of Computer Science and Engineering, The Chinese University of Hong Kong\\
    \textsuperscript{\rm 2}IBM Research \\
    \{zyhe, yjyang, qxu, tyho\}@cse.cuhk.edu.hk,  pin-yu.chen@ibm.com
}

\usepackage{bibentry}

\begin{document} 

\maketitle
\begin{abstract}
Deep Neural Networks (DNNs) have achieved excellent performance in various fields. However, DNNs' vulnerability to Adversarial Examples (AE) hinders their deployments to safety-critical applications.
This paper presents a novel AE detection framework, named \textbf{\beyond}, for trustworthy predictions.
\beyond performs the detection by distinguishing the AE's abnormal relation with its augmented versions, i.e. neighbors, from two prospects: representation similarity and label consistency. 
An off-the-shelf Self-Supervised Learning (SSL) model is used to extract the representation and predict the label for its highly informative representation capacity compared to supervised learning models. For clean samples, their representations and predictions are closely consistent with their neighbors, whereas those of AEs differ greatly.  
Furthermore, we explain this observation and show that by leveraging this discrepancy \beyond can effectively detect AEs.
We develop a rigorous justification for the effectiveness of \beyond.
Furthermore, as a plug-and-play model, \beyond can easily cooperate with the Adversarial Trained Classifier (ATC), achieving the state-of-the-art (SOTA) robustness accuracy. 
Experimental results show that \beyond outperforms baselines by a large margin, especially under adaptive attacks.
Empowered by the robust relation net built on SSL, we found that \beyond outperforms baselines in terms of both detection ability and speed.
\textit{Our code will be publicly available.} 
\end{abstract}

\input{./sections/1_introduction.tex}
\input{./sections/2_relatedworks.tex}
\input{./sections/3_method.tex}

\input{./sections/3_5_theoretical_justification.tex}
\input{./sections/5_evaluation.tex}

\input{./sections/6_adaptive_attack.tex}

\input{./sections/7_discussion.tex}
\input{./sections/8_conclusion.tex}
\bibliography{LaTeX/main.bib}

\end{document}

%% file: sections/1_introduction.tex
\section{Introduction}
\label{sec:introduction}

\begin{figure*}[ht]
    \centering
    \includegraphics[scale=0.65]{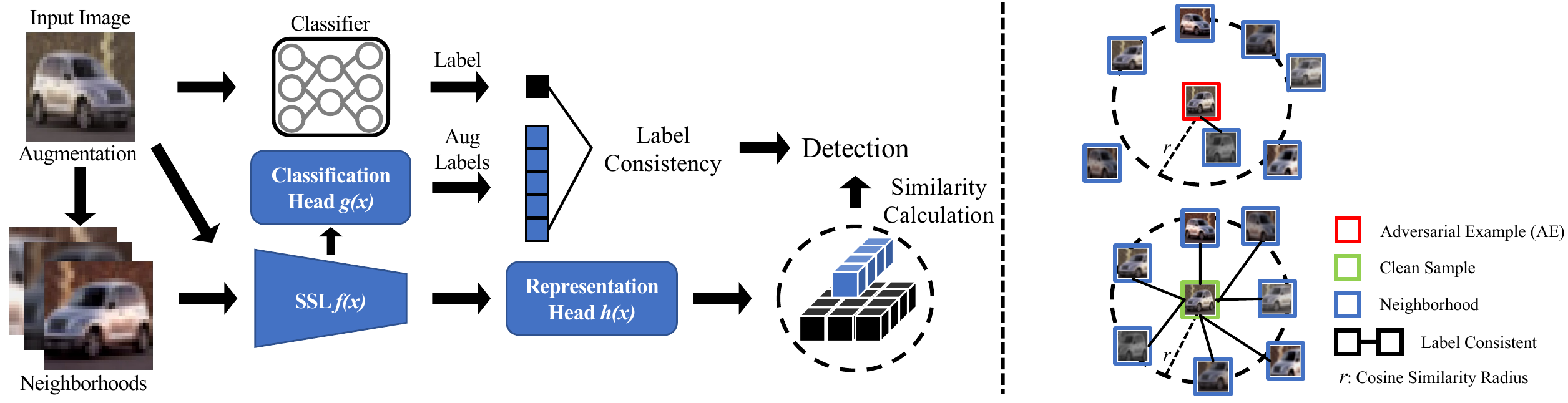}
    \captionof{figure}{Pipeline of the proposed \textbf{\beyond} framework. First we augment the input image to obtain a bunch of its neighbors. Then, we perform the label consistency detection mechanism on the classifier's prediction of the input image and that of neighbors predicted by SSL's classification head. Meanwhile, the representation similarity mechanism employs \textit{cosine distance} to measure the similarity among the input image and its neighbors (left). The input image with poor label consistency or representation similarity is flagged as AE (right).}
    \label{fig:framework}
\end{figure*}


Deep Neural Networks (DNNs) have been widely adopted in many fields due to their superior performance. However, DNNs are vulnerable to Adversarial Examples (AEs), which can easily fool DNNs by adding some imperceptible adversarial perturbations.
This vulnerability prevents DNN from being deployed in safety-critical applications such as autonomous driving~\cite{autodrive} and disease diagnosis~\cite{disease}, where incorrect predictions can lead to catastrophic economic and even loss of life.

Existing defensive countermeasures can be roughly categorized as: adversarial training, input purification~\cite{purify2}, and AE detection~\cite{fs}. Adversarial training is known as the most effective defense technique~\cite{autoattack}, but it brings about degradation of accuracy and additional training cost, which are unacceptable in some application scenarios. In contrast, input transformation techniques without high training / deployment costs, but their defensive ability is limited, i.e. easily defeated by adaptive attacks~\cite{autoattack}.

Recently, a large number of AE detection methods have been proposed. 
Some methods detect AE by interrogating the abnormal relationship between AE and other samples.
For example, Deep k-Nearest Neighbors (DkNN) \cite{dknn} compares the DNN-extracted features of the input image with those of its k nearest neighbors layer by layer to identify AE, leading to a high inference cost.  Instead of comparing all features,  Latent Neighborhood Graph (LNG) \cite{lng} employs a Graph Neural Network to make the comparison on a neighborhood graph, whose nodes are pre-stored embeddings of AEs together with those of the clean ones extracted by  DNN, and the edges are built according to distances between the input node and every reference node.  
Though more efficient than DkNN, LNG suffers from some weaknesses: some AE samples are required to build the graph the detection performance relays on the reference AEs and cannot effectively generalize to unseen attacks.
More importantly, both DkNN and LNG can be bypassed by adaptive attacks, in which the adversary has full knowledge of the detection strategy. 

We observe that one cause for adversarial vulnerability is the lack of feature invariance~\cite{feauture_invar}, i.e., small perturbations may lead to undesired large changes in features or even predicted labels. Self-Supervised Learning (SSL)~\cite{simsiam} models learn data representation consistency under different data augmentations, which intuitively can mitigate the issue of lacking feature invariance and thus improve adversarial robustness.
As an illustration, we visualize the SSL-extracted representation of the clean sample, AE and that of their corresponding augmentations in Fig.~\ref{fig:framework} (right). 
We can observe that the clean sample has closer ties with its neighbors, reflected by the higher label consistency and representation similarity. However, the AE representation stays quite far away from its neighbors, and there is a wide divergence in the predicted labels.

Inspired by the above observations, we propose a novel AE detection framework, named \textbf{\underline{BE} \underline{Y}our \underline{O}wn \underline{N}eighborhoo\underline{D} (\beyond)}. 
The contributions of this work are summarized as follows:
\begin{itemize}[leftmargin=*, noitemsep, topsep=0pt]
    \item We propose \beyond, a novel AE detection framework, which takes advantage of SSL model's robust representation capacity to identify AE by referring to its neighbors. 
    To our best knowledge, \beyond is the first work that leverages SSL model for AE detection without prior knowledge of adversarial attacks or AEs. 
    \item We develop a rigorous justification for the effectiveness of \beyond, and we derive an indicator to evaluate the validity of the candidate augmentation. 
    \item \beyond can defend effectively against adaptive attacks. 
    To defeat the two detection mechanisms: label consistency and representation similarity simultaneously, attackers have to optimize two objectives with contradictory directions, resulting in gradients canceling each other out.
    \item As a plug-and-play method, \beyond can be applied directly to any image classifier without compromising accuracy or additional retraining costs.

\end{itemize}
Experimental results show that \beyond outperforms baselines by a large margin, especially under adaptive attacks.
Empowered by the robust relation net built on SSL, we found \beyond outperforms baselines in terms of both detection ability and implementation costs.

%% file: sections/2_relatedworks.tex
\section{Related Works}
\label{sec:background}

\cite{intriguing} first discover that an adversary could maximize the prediction error of the network by adding some imperceptible perturbation, $\delta$, which is typically bounded by a perturbation budget, $\epsilon$, under a $L_p$-norm, e.g., $L_{\infty}$, $L_{2}$.
Project Gradient Descent (PGD) proposed by \cite{pgd} is one of the most powerful iterative attacks.
PGD motivates various gradient-based attacks such as AutoAttack \cite{autoattack} and Orthogonal-PGD \cite{opgd}, which can break many SOTA AE defenses~\cite{eval_robust}. 
Another widely adopted adversarial attack is C\&W~\cite{cw}. Compared to the norm-bounded PGD attack, C\&W conducts
AEs with a high attack success rate by formulating the adversarial attack problem as an optimization problem.  

Existing defense techniques focus either on robust prediction or detection.
The most effective way to achieve robust prediction is adversarial training~\cite{trade, sigmoid}, and the use of nearest neighbors is a common approach to detecting AEs. kNN~\cite{knn} and DkNN~\cite{dknn} discriminate AEs by checking the label consistency of each layer’s neighborhoods. \cite{lid} define Local Intrinsic Dimensionality (LID) to characterize the properties of AEs and use a simple k-NN classifier to detect AEs. LNG~\cite{lng} searches for the nearest samples in the reference data and constructs a graph, further training a specialized GNN to detect AEs.
Although these nearest-neighbor-based methods achieve competitive detection performance, all rely on external AEs for training detectors or searching thresholds, resulting in defeat against unseen attacks.

Recent studies have shown that SSL can improve adversarial robustness as SSL models are label-independent and insensitive to transformations \cite{ssl_robust}. An intuitive idea is to combine adversarial training and SSL \cite{ssl_at1} \cite{ssl_at2}, which remain computationally expensive and not robust to adaptive attacks. \cite{purify1} and \cite{purify2} find that the auxiliary SSL task can be used to purify AEs, which are shown to be robust to adaptive attacks.
However, \cite{eval_robust} show that these purification methods can be broken by stronger adaptive attacks. 

%% file: sections/3_method.tex
\section{Proposed Method}
\label{sec:methodology}

\subsection{Method Overview}
\label{sec:overview}
\subsubsection{Components.}
\beyond consists of three components: a SSL feature extractor $f(\cdot)$, a classification head $g(\cdot)$, and a representation head $h(\cdot)$, as shown in Fig.~\ref{fig:framework} (left). 
Specifically, the SSL feature extractor is a Convolutional Neural Network (CNN), pre-trained by specially designed loss, e.g. contrastive loss, without supervision~\footnote{Here, we employ the SimSiam~\cite{simsiam} as the SSL feature-extractor for its decent performance.}. A Fully-Connected layer (FC) acts as the classification head $g(\cdot)$, trained by freezing the $f(\cdot)$. The $g(\cdot)$ performs on the input image's neighbors for label consistency detection. The representation head $h(\cdot)$ consisting of three FCs, encodes the output of $f(\cdot)$ to an embedding, i.e. representation. We operate the representation similarity detection between the input image and its neighbors. 
\subsubsection{Core idea.}
Our approach relies on robust relationships between the input and its neighbors for the detection of AE. The key idea is that adversaries may easily attack one sample's representation to another submanifold, but it is difficult to totally shift that of all its neighbors. 
We employ the SSL model to capture such relationships, since it is trained to project input and its augmentations (neighbors) to the same submanifold~\cite{simsiam}. 
\subsubsection{Selection of neighbor number.}
Obviously, the larger the number of neighbors, the more stable the relationship between them, but this may increase the overhead. We choose 50 neighbors for \beyond, since larger neighbors no longer significantly enhance performance, as shown in Fig.~\ref{fig:k_pgd}.  
\subsubsection{Workflow.}
Fig.~\ref{fig:framework} shows the workflow of the proposed \beyond. When input comes, we first transform it into 50 augmentations, i.e. 50 neighbors. Note that \beyond is not based on random data augmentation. Next, the input along with its 50 neighbors are fed to SSL feature extractor $f(\cdot)$ and then the classification head $g(\cdot)$ and the representation head $h(\cdot)$, respectively. For the classification branch, $g(\cdot)$ outputs the predicted label for 50 neighbors. 
Later, the label consistency detection algorithm calculates the consistency level between the input label (predicted by the classifier) and 50 neighbor labels. When it comes to the representation branch, the 51 generated representations are sent to the representation similarity detection algorithm for AE detection. If the consistency of the label of a sample or its representation similarity is lower than a threshold, \beyond shall flag it AE. 

\begin{algorithm}[tbp]
\caption{\beyond detection algorithm}
\label{alg:beyond}
\small
\textbf{Input}: Input image $x$, target classifier $c(\cdot)$, SSL feature exactor $f(x)$, classification head $g(x)$, projector head $h(x)$, label consistency threshold $\mathcal{T}_{label}$, representation similarity threshold $\mathcal{T}_{rep}$, Augmentation $Aug$, neighbor indicator $i$, total neighbor $k$\\
\textbf{Output}: \texttt{reject / accept} \\
\DontPrintSemicolon
\begin{algorithmic}[1]
\STATE \textit{\textbf{Stage1: Collect labels and representations.}} \\
\STATE      $\ell_{cls}(x) = c(x)$ \\
\STATE \textbf{for} {$i$ in $k$} \textbf{do}
\STATE \ \ \ \ $\hat{x}_{i} = Aug(x)$
\STATE $\ell_{ssl}(\hat{x}_i) = f(g(\hat{x}_i))$;$r(x) = f(h(x))$; $r(\hat{x}_i) = f(h(\hat{x}_i))$
\STATE \textit{\textbf{Stage2: Label consistency detection mechanism.}} \\
\STATE \textbf{for} {$i$ in $k$} \textbf{do}
\STATE \ \ \ \  \textbf{if} {$\ell(\hat{x}_i)==\ell(x)$} \textbf{then} $\mathtt{Ind_{label}}+=1$
\STATE \textit{\textbf{Stage3: Representation similarity detection mechanism.}} \\
\STATE \textbf{for} {$i$ in $k$} \textbf{do}
\STATE \textbf{if} {$cos(r(x), r(\hat{x}_i))<\mathcal{T}_{cos}$} \textbf{then} $\mathtt{Ind}_{rep}+=1$
\STATE \textit{\textbf{Stage4: AE detection.}} \\
\STATE {\textbf{if }$\mathtt{Ind}_{label}<\mathcal{T}_{label}$ or $\mathtt{Ind}_{rep}<\mathcal{T}_{rep}$ \textbf{then} \texttt{reject}}
\STATE {\textbf{else} \texttt{accept}}
\end{algorithmic}
\end{algorithm}
\subsection{Detection Algorithms}
For enhanced AE detection capability, \beyond adopted two detection mechanisms: \textit{Label Consistency}, and \textit{Representation Similarity}. The detection performance of the two combined can exceed any of the individuals. More importantly, their contradictory optimization directions hinder adaptive attacks to bypass both of them simultaneously. 

\noindent \textbf{Label Consistency.} We compare the classifier prediction, $\ell_{cls}(x)$, on the input image, $x$, with the predictions of the SSL classification head, $\ell_{ssl}(\hat{x}_i), i=1\dots k$, where $\hat{x}_i$ denotes the $i$th neighbor,  $k$ is the total number of neighbors. If $\ell_{cls}(x)$ equals $\ell_{ssl_i}(\hat{x}_i)$, the label consistency increases by one, $\mathtt{Ind_{Label}+=1}$. Once the final label consistency is less than the threshold, $\mathtt{Ind_{Label}}<\mathcal{T}_{label}$, the \textit{Label Consistency} flags it as AE. We summarize the label consistency detection mechanism in Algorithm.~\ref{alg:beyond}.

\noindent \textbf{Representation Similarity.}  We employ the \textit{cosine distance} as a metric to calculate the similarity between the representation of input sample $r(x)$ and that of its neighbors, $r(\hat{x}_i), i=1,...,k$. Once the similarity, $ -cos(r(x),r(\hat{x}_i))$, is higher than a certain value, representation similarity increases by 1, $\mathtt{Ind_{Rep}+=1}$. If the final representation similarity is less than a threshold, $\mathtt{Ind_{ReP}}<\mathcal{T}_{rep}$, the \textit{representation similarity} flag the sample as an AE. Algorithm.~\ref{alg:beyond} concludes the representation similarity detection mechanism.   

Note that, we select the thresholds, i.e. $\mathcal{T}_{label}$, $\mathcal{T}_{rep}$, by fixing the False Positive Rate (FPR)@5\%, which can be determined only by clean samples, and the implementation of our method needs no prior knowledge about AE.

\subsection{Resistance to Adaptive Attacks}
\label{sec:analysis_ada}
Attackers may design adaptive attacks to bypass \beyond, if attackers know both the classifier and the detection strategy. \beyond apply augmentations on the input, weakening adversarial perturbations. As a result, to fool SSL's classification results on neighbors, i.e. bypass label consistency detection, large perturbations are needed. However, the added perturbation should be small to bypass the representation similarity detection, since a large perturbation can alter the representation significantly. Therefore, to attack \beyond, attackers have to optimize two objectives that have contradictory directions, resulting in gradients canceling each other out.

%% file: sections/3_5_theoretical_justification.tex
\section{Theoretical Justification}
\label{sec:theroetical}
Our proposed method is based on the observation that the similarity between AE and its neighbors (augmentations) is significantly smaller than that of the clean sample in the feature space of the SSL model.
This section provides theoretical support for the above observation. 
For ease of exposition, our analysis is built on a first-order approximation of nonlinear neural networks, employs $\|\cdot\|_2$ to represent similarity, and considers the linear data augmentation.

\subsubsection{Theoretical analysis.} Given a clean sample $x$, we receive its feature $f(x)$ lying in the feature space spanned by the SSL model.
We assume that benign perturbation, i.e. random noise, $\hat{\delta}$, with bounded budgets causes minor variation, $\hat{\varepsilon}$, on the feature space, as described in Eq.~\ref{eq:benignperturb}:
\begin{equation}
\label{eq:benignperturb}
    f(x+\hat{\delta}) = f(x) + \nabla f(x)\hat{\delta} = f(x) + \hat{\varepsilon},
\end{equation}
where $\|\hat{\varepsilon}\|_{2}$ is constrained to be within a radius $r$.
In contrast, when it comes to AE, $x_{adv}$, the adversarial perturbation, $\delta$, can cause considerable change, due to its maliciousness, that is, it causes misclassification and transferability~\cite{papernot2016transferability,demontis2019adversarial,liu2021self}, as formulated in Eq.~\ref{eq:aeperturb}.
\begin{equation}
\label{eq:aeperturb}
    f(x_{adv}) = f(x+ \delta) = f(x) + \nabla f(x)\delta = f(x) + \varepsilon,
\end{equation}
where $\|\varepsilon\|_2$ is significantly larger than $\|\hat{\varepsilon}\|_{2}$ formally,
    $\lim_{\hat{\varepsilon} \to 0} \frac{\varepsilon}{ \hat{\varepsilon}}=\infty$.
SSL model is trained to generate close representations for an input $x$ and its augmentation $x_{aug} = Wx$~\cite{jaiswal2020survey,hendrycks2019using}, where $W\in \mathbb{R}^{w \times h}$, $w$, $h$ denote the width and height of $x$, respectively. Based on this natural property of SSL,  we have:
\begin{small}
\begin{equation}
\label{eq:approx}
    f(Wx)=f(x)+o(\hat{\varepsilon}), \\
    \nabla f(Wx)=\nabla f(x)+o(\hat{\varepsilon}),
\end{equation}
\end{small} 

\noindent where $o(\hat{\varepsilon})$ is a high-order infinitesimal item of $\hat{\varepsilon}$. Moreover, according to Eq.~\ref{eq:benignperturb} and Eq.~\ref{eq:approx}, we can derive that:
\begin{small}
\begin{equation}
\label{eq:aug_benign}
\begin{aligned}
    f(W(x+\hat{\delta}) &= f(Wx) + \nabla f(Wx)W\hat{\delta}\\
    &= f(x) + \nabla f(x)W\hat{\delta} +o(\hat{\varepsilon}).
\end{aligned}
\end{equation}
\end{small}

\noindent We let $\hat{\varepsilon}_{aug}=\nabla f(x)W\hat{\delta}$ and assume $\hat{\varepsilon}_{aug}$ and $\hat{\varepsilon}$ are infinitesimal isotropic, i.e. $\lim_{\hat{\varepsilon} \to 0} \frac{\hat{\varepsilon}_{aug}}{\hat{\varepsilon}}=c$, where $c$ is a constant. Therefore, we can rewrite Eq.~\ref{eq:aug_benign} as follows:
\begin{small}
\begin{equation}
    f(W(x+\hat{\delta})) = f(x) + c\cdot\hat{\varepsilon} + o(\hat{\varepsilon}).
\end{equation}
\end{small}

Our goal is to prove that \textit{distance (similarity) between AE and its neighbors can be significantly smaller (larger) than that of the clean sample in the space spanned by a SSL model}, which is equivalent to justify Eq.~\ref{eq:objective_2}  
\begin{small}
\begin{equation}
\label{eq:objective_2}
     \|f(x_{adv}) - f(W(x_{adv}))\|_{2}^{2} \geq \|\underbrace{f(x) -f(Wx)}_{\hat{\varepsilon}_{aug}=c\cdot\hat{\varepsilon}}\|_{2}^{2}.
\end{equation}
\end{small}

\noindent Expending the left-hand item in Eq.~\ref{eq:objective_2}, and defining $m=\nabla f(x)W\delta$, we can obtain the following.
\begin{small}
\begin{equation}
\begin{aligned}
    \label{eq:expend}
   & \|f(x_{adv}) - f(Wx_{adv})\|_{2}^{2} =  \|f(x+\delta) - f(W(x+\delta))\|_{2}^{2}\\ 
   & = \|f(x)+\nabla f(x)\delta - f(Wx)-\nabla f(Wx)W\delta\|_{2}^{2}\\ 
    & =\|\varepsilon -\nabla f(x)W\delta - o(\varepsilon)\|_{2}^2 = \|\varepsilon\|_{2}^2 + \|m\|_{2}^2 - 2|\langle \varepsilon, m\rangle|+o(\varepsilon)
\end{aligned}
\end{equation}
\end{small}

\noindent As mentioned in the prior literature~\cite{mikolajczyk2018data,zeng2020data,raff2019barrage}, augmentations can effectively weaken adversarial perturbation $\delta$. Therefore, we assume that the influence caused by $W\delta$ is weaker than $\delta$ but stronger than the benign perturbation, $\hat{\delta}$. Formally, we have:
\begin{small}
\begin{equation}
\label{eq:aug_weaken}
    \|\underbrace{\nabla f(x)\delta}_{\varepsilon}\|_{2} \textgreater \|\underbrace{\nabla f(x)W\delta}_{m} \|_{2} \textgreater \|\underbrace{\nabla f(x)W\hat{\delta}}_{\hat{\varepsilon}_{aug}=c\cdot\hat{\varepsilon}}\|_{2}.
\end{equation}
\end{small}

\noindent According to \textit{Cauchy–Schwarz inequality}~\cite{bhatia1995cauchy}, we have the following chain of inequalities obtained by taking Eq.~\ref{eq:aug_weaken} into Eq.~\ref{eq:expend}:
\begin{small}
\begin{equation}
\label{eq:results}
\begin{aligned} 
    &\|\varepsilon\|_{2}^2 + \|m\|_{2}^2 - 2|\langle\varepsilon, m\rangle|+o(\varepsilon) \textgreater \\ 
    &\|\varepsilon\|_{2}^2 + \|m\|_{2}^2 - 2\|\varepsilon\|\cdot\| m\| = (\|\varepsilon\|_{2} - \|m\|_{2})^2,
\end{aligned}
\end{equation}
\end{small} 

\noindent where $ \|m\|\in{(\|\hat{\varepsilon}_{aug}\|,\|\varepsilon\|)}$ according to 
Eq.~\ref{eq:aug_weaken}.

Finally, from Eq.~\ref{eq:results} we observe that by applying proper data augmentation, the distance between AE and its neighbors in SSL's feature space$\|f(x_{adv})-f(Wx_{adv})\|_{2} = \left \| \|\varepsilon\|_{2} - \|m\|_{2}\right\|_2$ can be significantly larger than that of clean samples $\|f(x)-f(Wx)\|_2 = o(\hat{\varepsilon})$. The enlarged distance is upper bounded by $\|\varepsilon\|_{2} / \|\hat{\varepsilon}_{aug}\|_{2}$ times that of clean sample, which implies that the imperceptible perturbation $\delta$ in the image space can be significantly enlarged in SSL's feature space by referring to its neighbors.  
This exactly supports the design of \beyond as described in Section~\ref{sec:overview}.
In practice, we adopt various augmentations instead of a single type to generate multiple neighbors for AE detection, which reduces the randomness, resulting in more robust estimations.

\subsubsection{Select effective augmentations.} To better improve the effectiveness of \beyond, we analyze the conditions under which the augmentation can effectively weaken adversarial perturbation. 
Effective data augmentation makes the augmented label $y_{a u g}$ tend to the ground-truth label $y_{t r u e}$ and away from the adversarial label $y_{a d v}$:
\begin{equation}
    || y_{a u g}-y_{t r u e}||_{2} \leq ||y_{a u g}-y_{a d v}||_{2} \leq|| y_{a d v}-y_{t r u e}||_{2}.
\label{eq:distance}
\end{equation}
Since $y_{true}$ is the hard label (one-hot encoding), the range of $|| y_{a d v}-y_{t r u e} ||_{2}$ is $(\sqrt{2}/2, \sqrt{2})$. 
The distance is $\sqrt{2}$ when the item corresponding to $y_{a d v}$ is 1 in the logits of $y_{a d v}$, and $\sqrt{2}/2$ when the item corresponding to $y_{a d v}$ and $y_{t r u e}$ both occupy 1/2.
Given a SSL-based classifier, $C$, we have:
\begin{small}
\begin{equation}
\begin{aligned}
    &C(W(x+\delta)) = C(Wx)+\nabla C(Wx)W\delta \\
    &= y_{t r u e} + \nabla C(Wx)W\delta = y_{a u g}.
\end{aligned}
\end{equation}
\end{small}
Therefore, the distance between $y_{a u g}$ and $y_{t r u e}$ is: 
\begin{small}
\begin{equation}
\begin{aligned}
    &||y_{a u g} - y_{t r u e}||_2 = ||\nabla C(Wx)W\delta||_2 \\ &\leq ||\nabla C(Wx)W||_2 ||\delta||_2
    \leq ||\nabla C(Wx)W||_{2} \epsilon,
\end{aligned}
\end{equation}
\end{small}
where $||\delta||_2$ is bounded by $\epsilon$. Eq. \ref{eq:distance} always holds, then:
\begin{small}
\begin{equation}
\label{eq:aug_select}
    ||\nabla C(Wx)W||_2 \epsilon \leq \frac{\sqrt{2}}{2} \Rightarrow  ||\nabla C(Wx)W||_2 \leq \frac{\sqrt{2}}{{2\epsilon}}.
\end{equation}
\end{small}

In summary, augmentation can mitigate adversarial perturbation when it satisfies Eq. \ref{eq:aug_select}. We find \textit{color jitter} and \textit{rotation} are most effective in terms of Eq.~\ref{eq:aug_select} among all the tested data augmentations.
For space limitations, we leave the detailed comparison on data augmentations in Appendix.

%% file: sections/5_evaluation.tex
\section{Evaluation}
\label{sec:evaluation}

\subsection{Experimental Setting}

\textbf{Gray-box attack \& White-box attack.}
In the gray-box attack setting, the adversary has complete knowledge of the classifier, while the detection strategy is confidential. Whereas in an adaptive attack (white-box) setting, the adversary is aware of the detection strategy.

\noindent \textbf{Datasets \& Target models.}
We conduct experiments on three commonly adopted datasets including \cifar~\cite{cifar10}, \cifarh, and a more \imagenet~\cite{imagenet}. 
The details of the target models (classifiers), and the employed SSL models together with their original classification accuracy on clean samples are summarized in Tab.~\ref{tb:model} \footnote{The pre-trained SSL models for \cifar and \cifarh are from Solo-learn~\cite{solo}, and for \imagenet are from SimSiam~\cite{simsiam}.}.

\begin{table}[!h]
\centering
\scalebox{0.9}{
\setlength{\tabcolsep}{4mm}{
\begin{tabular}{c|c|cc}

\hline
\multirow{2}{*}{\textbf{Dataset}} & \multirow{2}{*}{\textbf{\begin{tabular}[c]{@{}c@{}}Classifier\\SSL \end{tabular}}} & \multicolumn{2}{c}{\textbf{Acc. on clean samples}$\uparrow$}                                                                                                               \\ \cline{3-4} 
                         &                           & \multicolumn{1}{c|}{\begin{tabular}[c]{@{}c@{}}Classifier\\      \end{tabular}} & \begin{tabular}[c]{@{}c@{}}SSL  \end{tabular} \\ \hline
\cifar                  & ResNet18                  & \multicolumn{1}{c|}{91.53\%}                                                     & 90.74\%                                                  \\ \hline
\cifarh                  & ResNet18                  & \multicolumn{1}{c|}{75.34\%}                                                     & 66.04\%                                                  \\ \hline
\imagenet                 & ResNet50                  & \multicolumn{1}{c|}{80.86\%}                                                     & 68.30\%                                                  \\ \hline
\end{tabular}}}
\caption{Information of datasets and models.}
\label{tb:model}
\end{table}

\noindent \textbf{Attacks.}
Gray-box evaluations are conducted on FGSM, PGD, C\&W, and AutoAttack. AutoAttack includes APGD, APGD-T, FAB-T, and Square~\cite{square}, where APGD-T and FAB-T are targeted attacks and Square is a black-box attack. As for adaptive attacks, we employed the most adopted PGD and Orthogonal-PGD, which is a recent adaptive attack designed for AE detectors.   

\noindent \textbf{Metrics.} Following previous works~\cite{Yang2022WhatYS,lng} we employ 
TPR@FPR, ROC curve \& AUC, and robust accuracy (RA) as evaluation metrics. 
\begin{itemize}
    \item \textbf{TPR@FPR:} TPR@FPR indicates the true positive rate (TPR) at a false positive rate (FPR) $\leq n$\%, which is used as the primary metric to assess detector performance.
    \item \textbf{ROC curve \& AUC:} Receiver Operating Characteristic (ROC) curves describe the impact of various thresholds on detection performance, and the Area Under the Curve (AUC) is an overall indicator of the ROC curve.
    \item \textbf{Robust Accuracy (RA)}: We employ RA as an evaluation metric, which can reflect the  overall system performance against adaptive attacks by considering both the classifier and the detector.
\end{itemize}

\noindent \textbf{Baselines.}
We choose five detection-based defense methods as baselines: kNN, DkNN, LID, \cite{hu} and LNG, which also consider the relationship between the input and its neighbors to some extent. 

\begin{table}[tpb]
\centering
\scalebox{0.9}{
\setlength{\tabcolsep}{3mm}{
\begin{tabular}{c|ccc}
\hline
\multirow{1}{*}{\textbf{Dataset}} & \multicolumn{1}{c|}{\multirow{1}{*}{\textbf{\cifar}}} & \multicolumn{1}{c|}{\multirow{1}{*}{\textbf{\cifarh}}} & \multirow{1}{*}{\textbf{\imagenet}} \\\hline
\textbf{Attack}                 & \multicolumn{3}{c}{\textbf{TPR@FPR5}\%$\uparrow$}                                                                                             \\ \hline
FGSM                     & \multicolumn{1}{c|}{86.16\%}                  & \multicolumn{1}{c|}{89.80\%}                   & 61.05\%                   \\ \hline
PGD                      & \multicolumn{1}{c|}{82.80\%}                  & \multicolumn{1}{c|}{85.90\%}                   & 89.80\%                   \\ \hline
C\&W                       & \multicolumn{1}{c|}{91.48\%}                  & \multicolumn{1}{c|}{91.96\%}                   & 76.69\%                   \\ \hline
AutoAttack               & \multicolumn{1}{c|}{93.42\%}                  & \multicolumn{1}{c|}{90.90\%}                   & 84.25\%                   \\ \hline
\end{tabular}}}
\caption{TPR@FPR 5\% of \beyond against Gray-box Attacks. All attacks are performed under $L_\infty=8/255$.} 
\label{tb:tpr2fpr}
\end{table}

\subsection{Defending Gray-box Attacks}
Tab. \ref{tb:tpr2fpr} reports TPR@FPR5\% to show the AE detection performance of \beyond. It can be observed that \beyond maintains a high detection performance on various attacks and datasets, which is attributed to our detection mechanism. Combining label consistency and representation similarity, \beyond identifies AEs without reference AE set.
For more results of TPR@FPR3\%, please refer to Appendix.

\begin{figure}[tpb]
    \centering
    \includegraphics[scale=0.35]{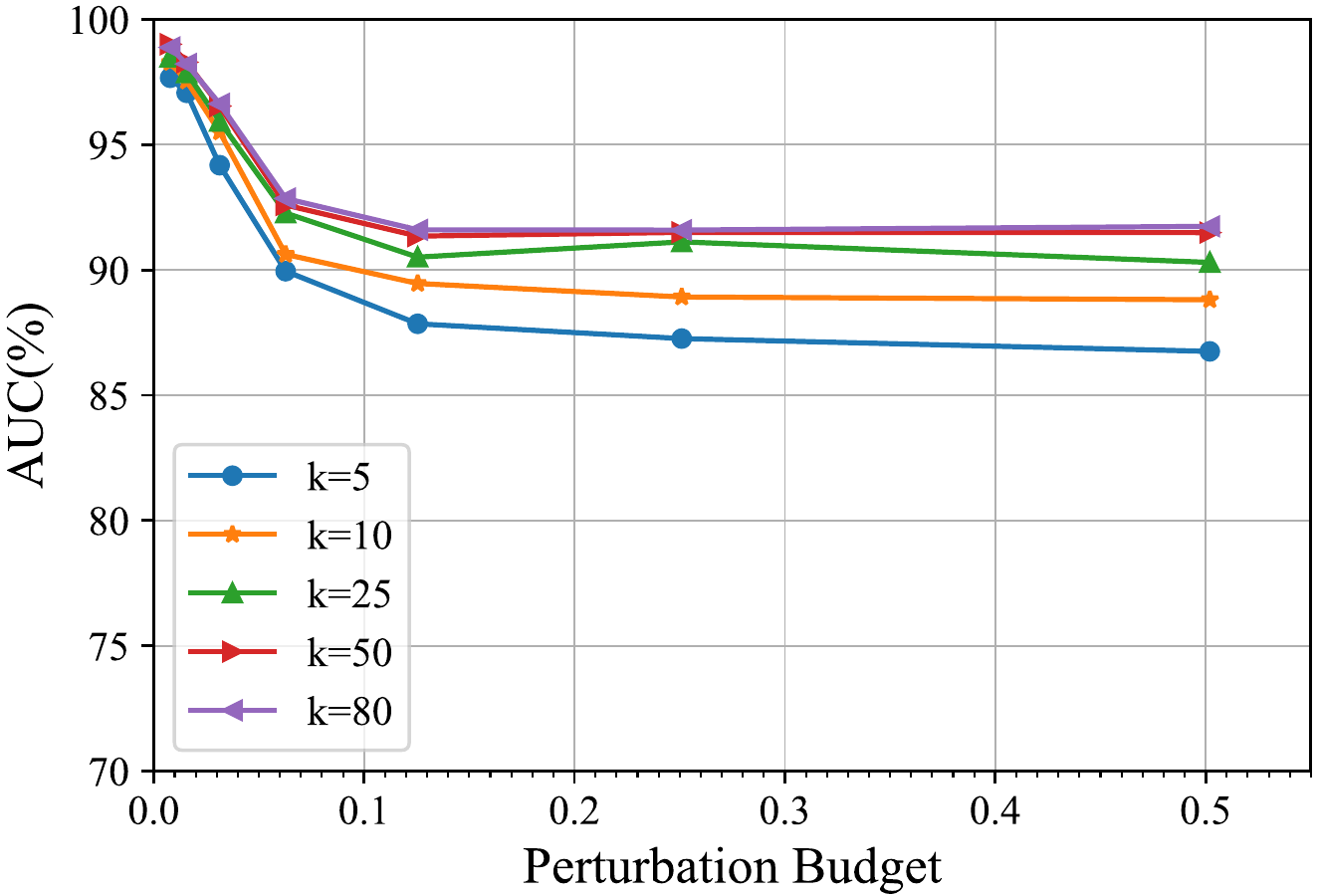}
    \caption{Detection performance with different number of neighbors on \cifar.}
    \label{fig:k_pgd}
\end{figure}

Fig. \ref{fig:k_pgd} shows the effect of the number of neighbors on the detection performance against PGD with different perturbation budgets. It can be observed that the detection performance with a large number of neighbors is better, but the performance is not significantly improved when the number of neighbors exceeds 50. 
In addition, previous methods argue that the detection of small perturbations is challenging~\cite{lng}. However, it's shown that \beyond has better detection performance for small perturbations, and maintains a high level of detection performance for large perturbations. This is because previous distance-measure-based methods are not sensitive to small perturbations, which \beyond avoids by combining differences in high-dimensional representations and semantics (label consistency).

\begin{table}[tpb]
\centering
\scalebox{0.8}{
\setlength{\tabcolsep}{1mm}{
\begin{tabular}{c|cc|cc}
\hline
\multirow{2}{*}{\textbf{Model}}  & \multicolumn{2}{c|}{\textbf{Acc. on clean samples}}                  & \multicolumn{2}{c}{\textbf{RA}}                    \\ \cline{2-5} 
                        & \multicolumn{1}{c|}{\textbf{ATC}}     & \textbf{ATC+\beyond} & \multicolumn{1}{c|}{\textbf{ATC}}     & \textbf{ATC+\beyond} \\ \hline
R2021Fixing70   & \multicolumn{1}{c|}{92.23\%} & \textbf{92.83\%}    & \multicolumn{1}{c|}{66.20\%} & \textbf{84.40\%}    \\ \hline
G2021Improving70  & \multicolumn{1}{c|}{88.74\%} & \textbf{90.81\%}    & \multicolumn{1}{c|}{64.10\%} & \textbf{81.50\%}    \\ \hline
G2020Uncovering70 & \multicolumn{1}{c|}{91.10\%} & \textbf{91.79\%}    & \multicolumn{1}{c|}{64.70\%} & \textbf{83.80\%}    \\ \hline
R2021Fixing106  & \multicolumn{1}{c|}{88.50\%} & \textbf{90.51\%}    & \multicolumn{1}{c|}{62.20\%} & \textbf{81.30\%}    \\ \hline
\end{tabular}}}
\caption{ATC+\beyond against AutoAttack on \cifar.}
\label{tb:atc}
\end{table}

\subsubsection{Combined With ATC}
As a plug-and-play approach, \beyond integrates well with existing Adversarial Trained Classifier (ATC) \footnote{All ATCs are sourced from RobustBench~\cite{robustbench}.}. Tab. \ref{tb:atc} shows the accuracy on clean samples and RA against AutoAttack of ATC combined with \beyond on \cifar.
As can be seen the addition of \beyond increases the robustness of ATC by a significant margin on both clean samples and AEs.

\begin{table*}[!ht]
\centering
\scalebox{0.9}{
\setlength{\tabcolsep}{4mm}{
\small
\begin{tabular}{c|cccc|ccccc}
\hline
\multirow{2}{*}{\textbf{Methods}} & \multicolumn{4}{c|}{\textbf{\textit{Unseen}: Attacks used in training are preclude from tests.}}                                                                                                              & \multicolumn{5}{c}{\textbf{\textit{Seen}: Attacks used in training are included in tests.}}                                                                                                                                               \\ \cline{2-10} 
                          & \multicolumn{1}{c|}{\textbf{FGSM}}             & \multicolumn{1}{c|}{\textbf{PGD}}              & \multicolumn{1}{c|}{\textbf{AutoAttack}}       & \textbf{Square}           & \multicolumn{1}{c|}{\textbf{FGSM}}    & \multicolumn{1}{c|}{\textbf{PGD}}              & \multicolumn{1}{c|}{\textbf{CW}}               & \multicolumn{1}{c|}{\textbf{AutoAttack}}       & \textbf{Square}           \\ \hline
DkNN                      & \multicolumn{1}{c|}{61.50\%}          & \multicolumn{1}{c|}{51.18\%}          & \multicolumn{1}{c|}{52.11\%}          & 59.51\%          & \multicolumn{1}{c|}{61.50\%} & \multicolumn{1}{c|}{51.18\%}          & \multicolumn{1}{c|}{61.46\%}          & \multicolumn{1}{c|}{52.11\%}          & 59.21\%          \\ \hline
kNN                       & \multicolumn{1}{c|}{61.80\%}          & \multicolumn{1}{c|}{54.46\%}          & \multicolumn{1}{c|}{52.64\%}          & 73.39\%          & \multicolumn{1}{c|}{61.80\%} & \multicolumn{1}{c|}{54.46\%}          & \multicolumn{1}{c|}{62.25\%}          & \multicolumn{1}{c|}{52.64\%}          & 73.39\%          \\ \hline
LID                       & \multicolumn{1}{c|}{71.15\%}          & \multicolumn{1}{c|}{61.27\%}          & \multicolumn{1}{c|}{55.57\%}          & 66.11\%          & \multicolumn{1}{c|}{73.56\%} & \multicolumn{1}{c|}{67.95\%}          & \multicolumn{1}{c|}{55.60\%}          & \multicolumn{1}{c|}{56.25\%}          & 85.93\%          \\ \hline
Hu                        & \multicolumn{1}{c|}{84.44\%}          & \multicolumn{1}{c|}{58.55\%}          & \multicolumn{1}{c|}{53.54\%}          & \underline{\textit{95.83\%}}          & \multicolumn{1}{c|}{84.44\%} & \multicolumn{1}{c|}{58.55\%}          & \multicolumn{1}{c|}{\underline{\textit{90.99\%}}}          & \multicolumn{1}{c|}{53.54\%}          & 95.83\%          \\ \hline
LNG                       & \multicolumn{1}{c|}{\underline{\textit{98.51\%}}}          & \multicolumn{1}{c|}{\underline{\textit{63.14\%}}}          & \multicolumn{1}{c|}{\underline{\textit{58.47\%}}}          & 94.71\%          & \multicolumn{1}{c|}{\textbf{99.88\%}} & \multicolumn{1}{c|}{\underline{\textit{91.39\%}}} & \multicolumn{1}{c|}{89.74\%}          & \multicolumn{1}{c|}{\underline{\textit{84.03\%}}}          & \underline{\textit{98.82\%}}          \\ \hline
\textbf{\beyond}                    & \multicolumn{1}{c|}{\textbf{98.89\%}} & \multicolumn{1}{c|}{\textbf{99.29\%}} & \multicolumn{1}{c|}{\textbf{99.18\%}} & \textbf{99.29\%} & \multicolumn{1}{c|}{\underline{\textit{98.89\%}}} & \multicolumn{1}{c|}{\textbf{99.29\%}} & \multicolumn{1}{c|}{\textbf{99.20\%}} & \multicolumn{1}{c|}{\textbf{99.18\%}} & \textbf{99.29\%} \\ \hline
\end{tabular}}}
\caption{The AUC of Different Adversarial Detection Approaches on \cifar. To align with baselines, classifier: ResNet110, FGSM: $\epsilon=0.05$, PGD: $\epsilon=0.02$. Note that \textbf{\beyond needs no AE for training}, leading to the same value on both \textit{seen} and \textit{unseen} settings. The \textbf{bolded} values are the best performance, and the \underline{\textit{underlined italicized}} values are the second-best performance, the same below.}
\label{tb:compare}
\end{table*}

\subsubsection{Comparison with Existing Methods}
Tab. \ref{tb:compare} compares the AUC of \beyond with five AE detection methods: DkNN, kNN, LID, Hu et al., and LNG on \cifar. Since LID and LNG rely on reference AEs, we report detection performance on both seen and unseen attacks. In the seen attack setting, LID and LNG are trained with all types of attacks, while using only the C\&W attack in the unseen attack setting. Note that the detection performance for seen and unseen attacks is consistent for detection methods without AEs training.

Experimental results show that \beyond consistently outperforms SOTA AE detectors, and the performance advantage is particularly significant when detecting unseen attacks. This is because \beyond uses the augmentations of the input as its neighbors without relying on prior adversarial knowledge. And according to the conclusion in Sec. \ref{sec:theroetical}, adversarial perturbations impair label consistency and representation similarity, which enables \beyond to distinguish AEs from benign ones with high accuracy. For more comparative results on \imagenet, please refer to the Appendix.

%% file: sections/6_adaptive_attack.tex
\subsection{Defending Adaptive Attacks}
\label{sec:adaptive}
\subsubsection{Adaptive Objective Loss Function.}
\label{sec:design_ada}
Attackers can design adaptive attacks to bypass \beyond when the attacker knows all the parameters of the model and the detection strategy. 
To attack effectively, the adversary must deceive the target model while guaranteeing the label consistency and representation similarity of the SSL model. 
Since \beyond uses multiple augmentations, we consider the impact of all augmentations on label consistency and representation similarity during the attack, and conduct the following adaptive attack objective items as suggested by~\cite{eot}:
\begin{small}
\begin{equation}
\begin{aligned}
    Sim_{l} &= \frac{1}{k} \sum_{i=1}^{k} \mathcal{L}\left(C\left(W^i(x+\delta)  \right), y_t\right) \\
    Sim_{r} &= \frac{1}{k} \sum_{i=1}^{k}(\mathcal{S}(r(W^i(x+\delta)), r(x+\delta))), \\
\end{aligned}
\label{eq:analysis_1}
\end{equation}
\end{small}

\noindent where $\mathcal{S}$ represents cosine similarity, and adaptive adversaries perform attacks on the following objective function:
\begin{small}
\begin{equation}
    \underset{\delta}{min}~{\mathcal{L}_C(x+\delta, y_t) + Sim_l - \alpha \cdot Sim_{r}},
    \label{eq:adaptive}
\end{equation}
\end{small}
\noindent where $\mathcal{L}_C$ indicates classifier's loss function, $y_t$ is the targeted class, $\delta$ denotes the adversarial perturbation, and $\alpha$ refers to a hyperparameter~\footnote{Note that we employ cosine metric is negatively correlated the similarity, so that the $Sim_{r}$ item is preceded by a minus sign.}.
\begin{table}[tpb]
\centering
\scalebox{0.8}{
\setlength{\tabcolsep}{0.7mm}{
\begin{tabular}{c|c|c|c|c|c|c}
\hline
\textbf{$\alpha$} & \textbf{0} & \textbf{1} & \textbf{10} & \textbf{20} & \textbf{50} & \textbf{100} \\ \hline
\cifar        & 82.03\%    & 63.91\%     & 64.57\%      & 76.15\%      & 88.56\%      & 92.53\%       \\ \hline
\cifarh       & 90.58\%    & 88.49\%     & 91.61\%      & 93.10\%      & 94.05\%      & 94.37\%       \\ \hline
\end{tabular}}}
\caption{AUC for Adaptive Attack under different $\alpha$ (Eq.\ref{eq:adaptive}).}
\label{tb:ada_alpha}
\end{table}
The design of the adaptive attack in Eq.~\ref{eq:adaptive} includes a hyperparameter $\alpha$, which is a trade-off parameter between label consistency and representation similarity. 
Tab. \ref{tb:ada_alpha} shows the AUC of \beyond under different $\alpha$. As shown, when $\alpha=0$, i.e. the attacker only attacks the label consistency detection mechanism, the AUC score is still high, 
which proves that our approach is not based on the weak transferability of AEs. 
Moreover, adaptive attacks are strongest when $\alpha=1$, which is used for rest tests. 


\begin{figure}[tpb]
    \centering
    \subfloat[\cifar]{
	\includegraphics[width=0.23\textwidth]{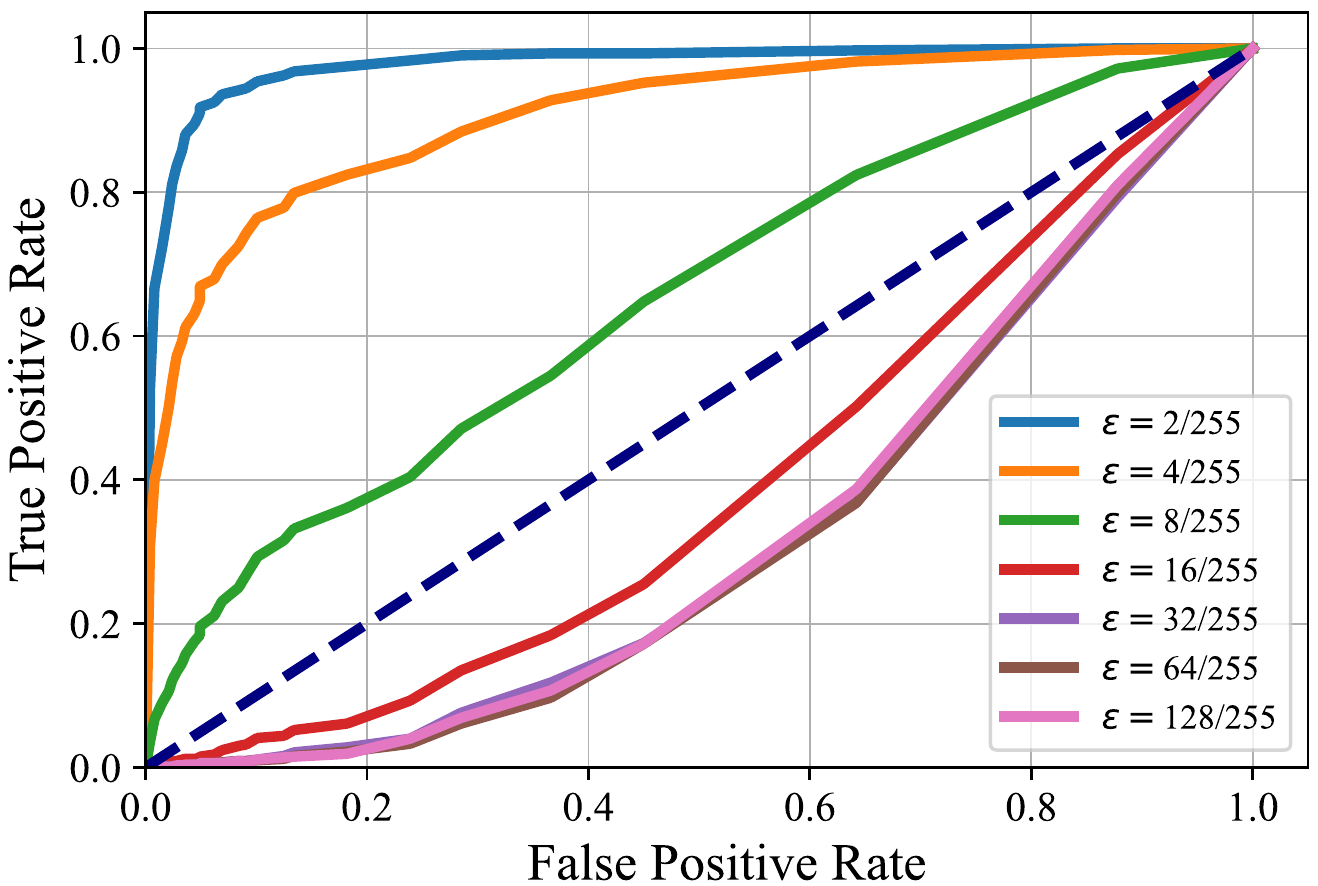}
	}%
    \subfloat[\cifarh]{
	\includegraphics[width=0.23\textwidth]{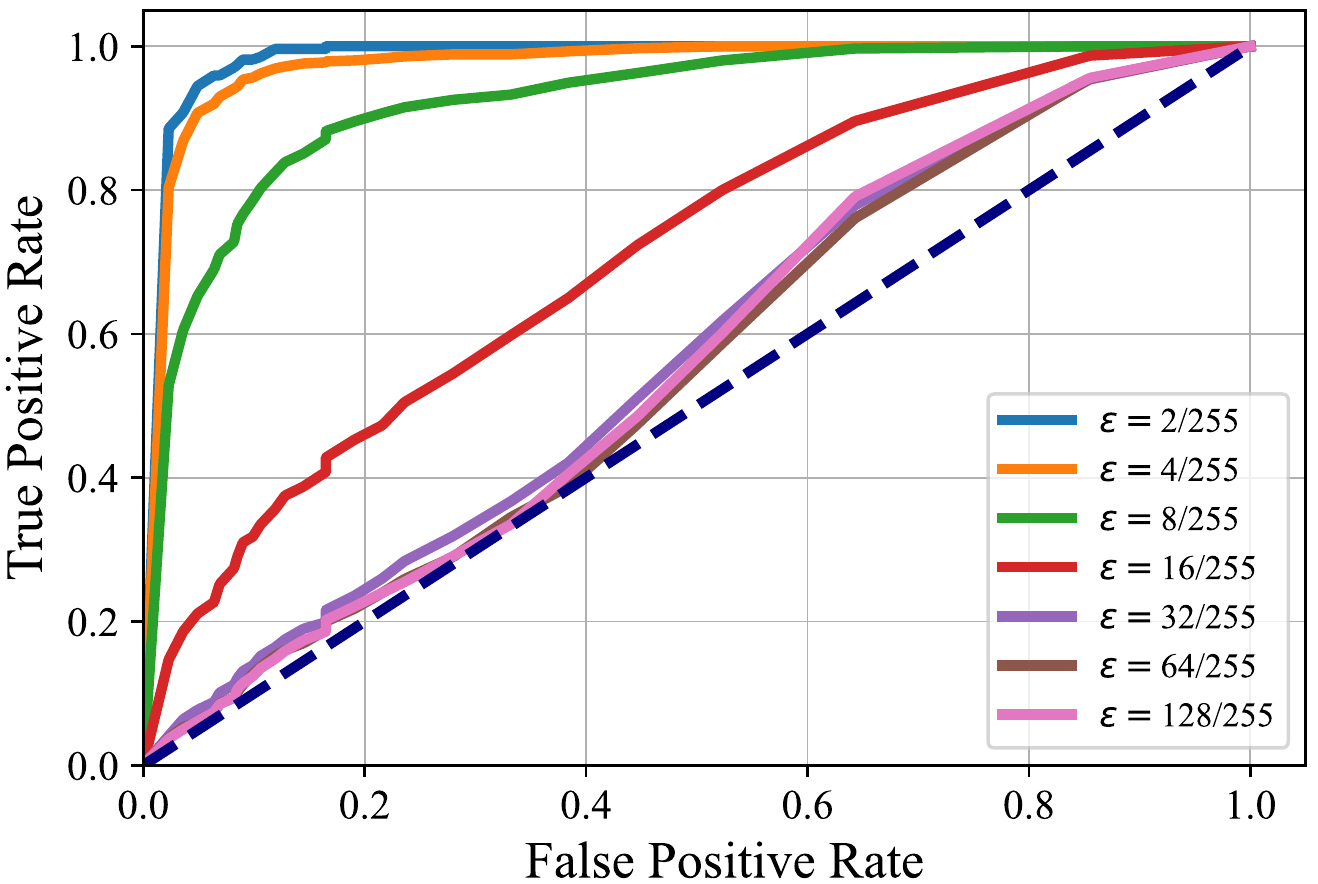}
	 }%
    \caption{ROC Curve of \beyond against adaptive attacks with different perturbation budgets.}
    \label{fig:roc_auc}
\end{figure}

\subsubsection{ROC Curve of Perturbation Budgets.} Fig. \ref{fig:roc_auc} summarizes the ROC curve varying with different perturbation budgets on \cifar and \cifarh. Our analysis regarding Fig. \ref{fig:roc_auc} is as follows: 1) \beyond can be bypassed when perturbations are large enough, due to large perturbations circumventing the transformation. This proves that \beyond is not gradient masking~\cite{obfuscated} and our adaptive attack design is effective. However, large perturbations are easier to perceive. 
2) When the perturbation is small, the detection performance of \beyond for adaptive attacks still maintains a high level, because small perturbations cannot guarantee both label consistency and representation similarity (as shown in Fig.~\ref{fig:ada_sim}). 
The above empirical conclusions are consistent with the intuitive analysis in Sec. \ref{sec:analysis_ada}.

\begin{table}[tpb]
\centering
\scalebox{0.8}{
\setlength{\tabcolsep}{0.2mm}{
\begin{tabular}{c|cc|cc}
\hline
\multirow{2}{*}{\textbf{Defense}} & \multicolumn{2}{c|}{\textbf{$L_{\infty}$=0.01}}                                                                                                       & \multicolumn{2}{c}{\textbf{$L_{\infty}$=8/255}}                                                                                                       \\ \cline{2-5} 
                                  & \multicolumn{1}{c|}{\begin{tabular}[c]{@{}c@{}}RA@FPR5\%\end{tabular}} & \begin{tabular}[c]{@{}c@{}}RA@FPR50\%\end{tabular} & \multicolumn{1}{c|}{\begin{tabular}[c]{@{}c@{}}RA@FPR5\%\end{tabular}} & \begin{tabular}[c]{@{}c@{}}RA@FPR50\%\end{tabular} \\ \hline
\textbf{\beyond }                           & \multicolumn{1}{c|}{{\underline{\textit{88.38\%}}}}                                    & {\underline{\textit{98.81\%}}}                                     & \multicolumn{1}{c|}{{\underline{\textit{13.80\%}}}}                                    & {\underline{\textit{48.20\%}}}                                     \\ \hline
\textbf{\beyond+ATC }                       & \multicolumn{1}{c|}{\textbf{96.30\%}}                                          & \textbf{99.30\%}                                           & \multicolumn{1}{c|}{\textbf{94.50\%}}                                          & \textbf{97.80\%}                                           \\ \hline
Trapdoor                          & \multicolumn{1}{c|}{0.00\%}                                                    & 7.00\%                                                     & \multicolumn{1}{c|}{0.00\%}                                                    & 8.00\%                                                     \\ \hline
DLA’20                            & \multicolumn{1}{c|}{62.60\%}                                                   & 83.70\%                                                    & \multicolumn{1}{c|}{0.00\%}                                                    & 28.20\%                                                    \\ \hline
SID’21                            & \multicolumn{1}{c|}{6.90\%}                                                    & 23.40\%                                                    & \multicolumn{1}{c|}{0.00\%}                                                    & 1.60\%                                                     \\ \hline
SPAM’19                           & \multicolumn{1}{c|}{1.20\%}                                                    & 46.00\%                                                    & \multicolumn{1}{c|}{0.00\%}                                                    & 38.00\%                                                    \\ \hline
\end{tabular}}}
\caption{Robust Accuracy under Orthogonal-PGD Attack.}

\label{tb:opgd}
\end{table}

\subsubsection{Performance against Orthogonal-PGD Adaptive Attacks}
Orthogonal-Projected Gradient Descent (Orthogonal-PGD) is a recently proposed AE detection benchmark. Orthogonal-PGD has two attack strategies: orthogonal strategy and selection strategy. Tab.~\ref{tb:opgd} shows \beyond outperforms the four baselines by a considerable margin in orthogonal strategy, especially under small perturbations. For the worst case, \beyond can still keep 13.8\% ($L_{\infty}=8/255$). Furthermore, incorporating ATC can significantly improve the detection performance of \beyond against large perturbation to 94.5\%. See Appendix for more selection strategy results.

\subsubsection{Trade-off between label consistency and Representation Similarity}
\begin{figure}[tpb]
    \centering
    \includegraphics[scale=0.35]{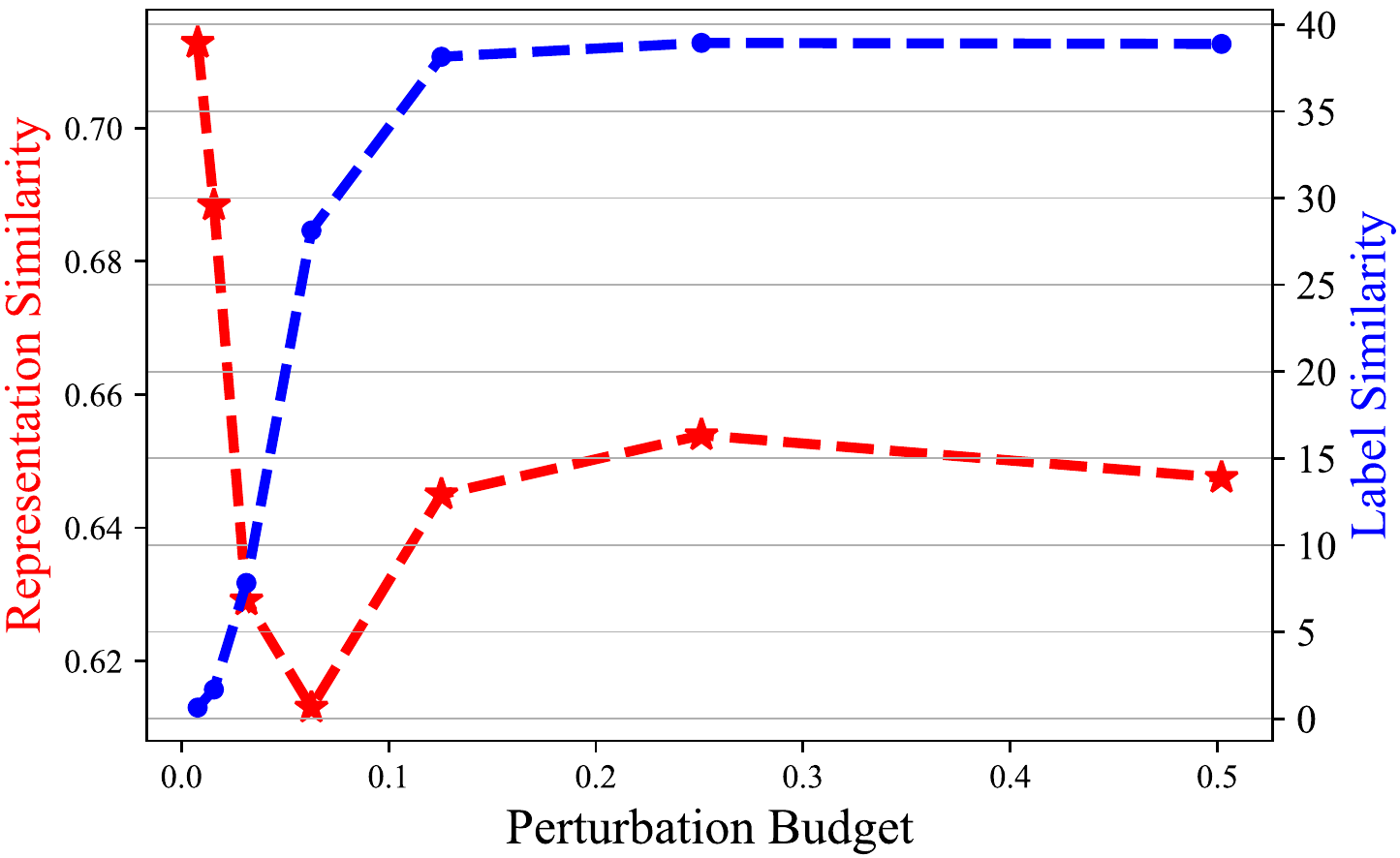}
    \caption{Trade-off between label consistency and representation similarity with perturbation budgets.}
    \label{fig:ada_sim}
\end{figure}

The previous intuitive analysis and empirical results have proved that there is a trade-off between label consistency and representation similarity. 
Fig. \ref{fig:ada_sim} shows the variation of label consistency and representation similarity with perturbation size on \cifarh. We can observe that label consistency and representation similarity respond differently to the perturbation size and they can be optimized simultaneously only when the perturbation is large enough.

\begin{figure}[tpb]
    \centering
    \includegraphics[scale=0.28]{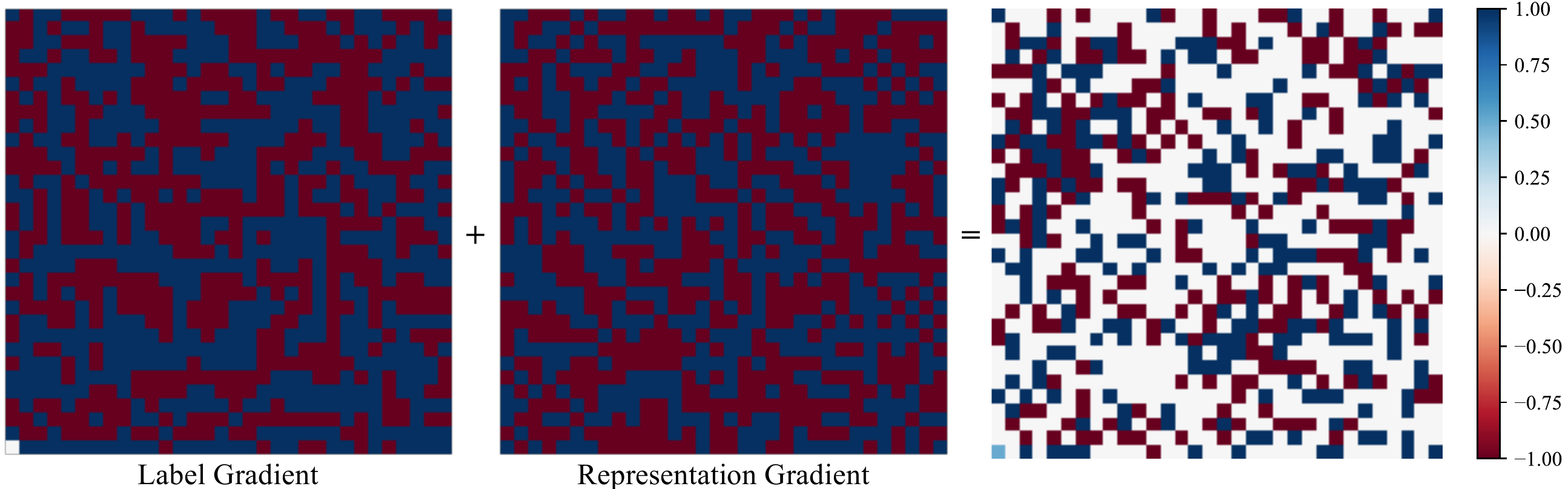}
    \caption{Gradient conflict between label consistency and representation similarity. The colored pixels represent the gradient direction, while the blank means gradient conflict.}
    \label{fig:grad_conflict}
\end{figure}

\begin{table}[hp]
\scalebox{0.8}{
\setlength{\tabcolsep}{1mm}{
\centering
\begin{tabular}{c|c|c|c|c|c|c|c}
\hline
\multirow{1}{*}{$\epsilon$} & \multirow{1}{*}{\textbf{2/255}} & \multirow{1}{*}{\textbf{4/255}} & \multirow{1}{*}{\textbf{8/255}} & \multirow{1}{*}{\textbf{16/255}} & \multirow{1}{*}{\textbf{32/255}} & \multirow{1}{*}{\textbf{64/255}} & \multirow{1}{*}{\textbf{128/255}} \\
                     \hline
s=0.002                & 55.92\%                 & 53.96\%                 & 52.32\%                 & 50.70\%                  & 50.73\%                  & 50.49\%                  & 50.36\%                   \\ \hline
s=0.01                 & 55.65\%                 & 52.91\%                 & 51.70\%                 & 48.53\%                  & 49.09\%                  & 49.09\%                  & 48.54\%                   \\ \hline
\end{tabular}}}
\caption{Gradient cancel ratio of perturbation budgets.}

\label{tb:grad}
\end{table}

The conflict between label consistency and representation similarity stems from their different optimization direction. Fig. \ref{fig:grad_conflict} visualizes the gradients produced by optimizing label consistency and representation similarity on the input. It's shown that attacks on label consistency or representation similarity produce gradients that modify the input in a certain direction, but optimizing for both leads to conflicting gradients. Tab. \ref{tb:grad} counts the gradient conflict rates under a range of perturbation budgets with different step sizes. We can discover that large perturbations result in reduced gradient conflict rates, which validates our analysis again.

\subsection{Ablation Study}



\begin{figure}[tbp]
    \subfloat[\cifar]{
	\includegraphics[width=0.23\textwidth]{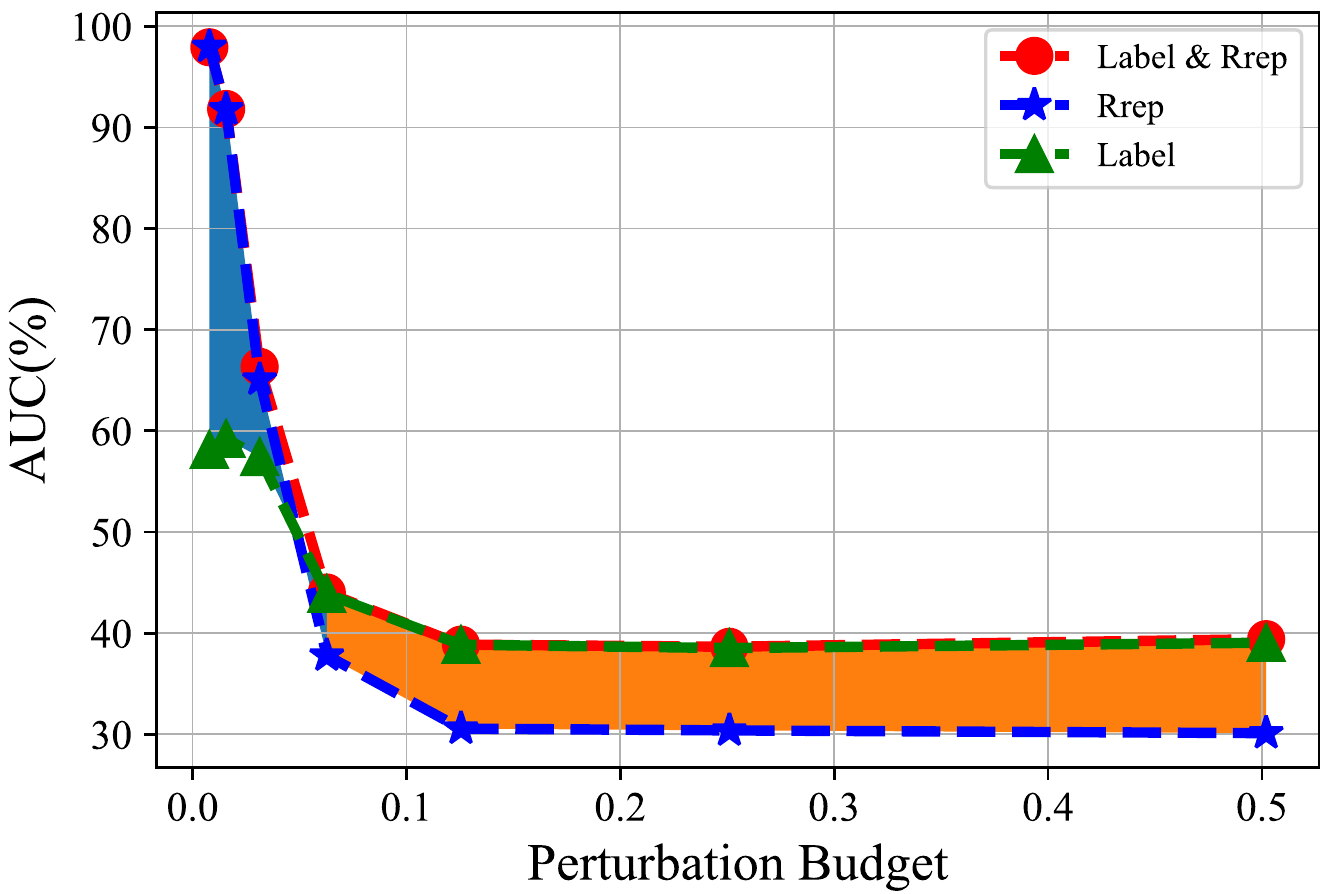}
	}%
    \subfloat[\cifarh]{
	\includegraphics[width=0.23\textwidth]{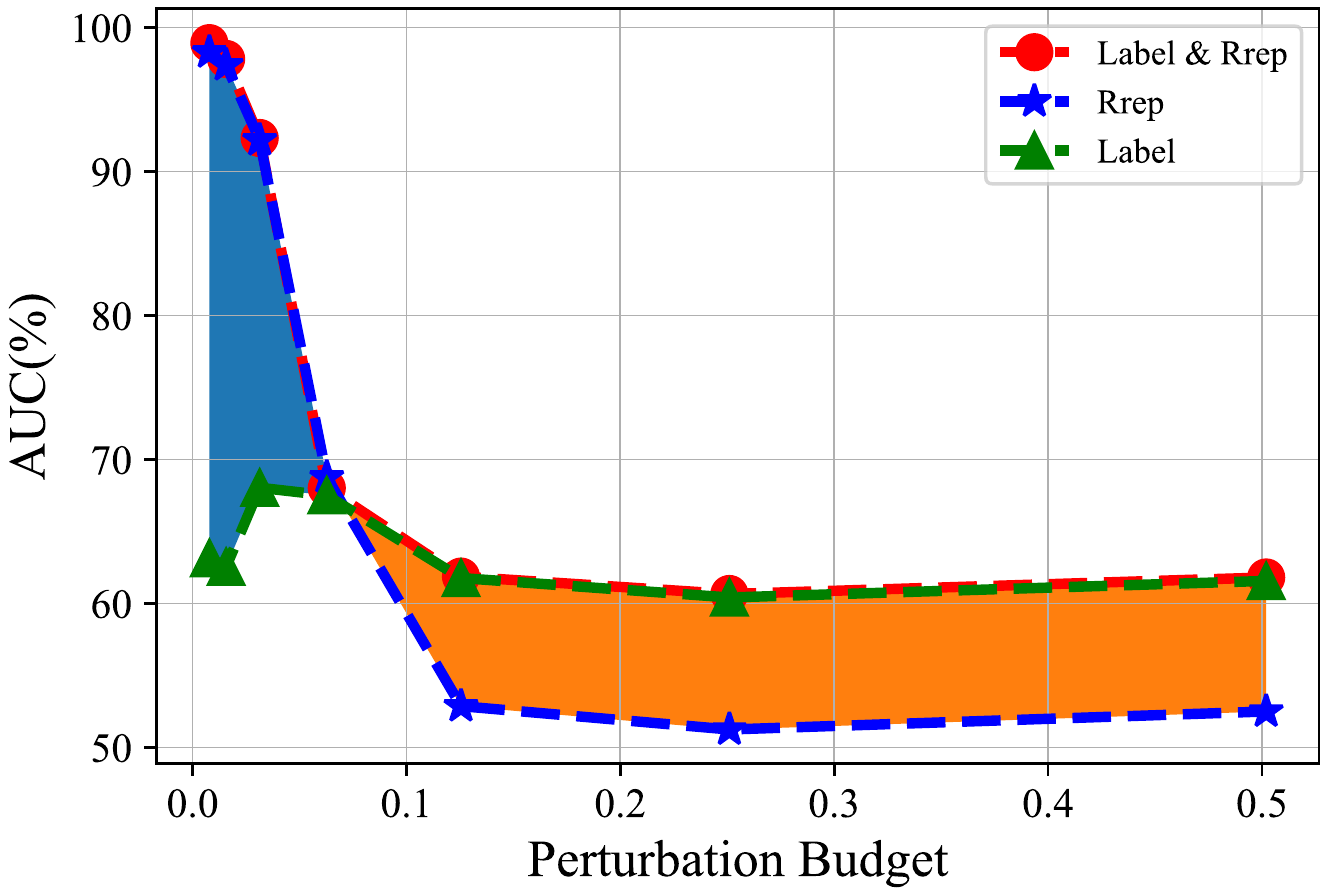}
	 }%
    \caption{Ablation study on \cifar and \cifarh.}
    \label{fig:ablation}
\end{figure}

We perform an ablation study on label consistency and representation similarity in \cifar and \cifarh, and analyze their detection performance (AUC) under a wide perturbation budget in Fig. \ref{fig:ablation}. 
When the perturbation is small, the detection performance based on label consistency (blue line) is better than representation similarity (green line). With increasing perturbation, representation similarity is difficult to maintain, leading to higher performance of representation similarity-based detectors, as same turning points as in Fig. \ref{fig:ada_sim}. In summary, the label consistency and representation similarity have different sensitivities to perturbation, so their cooperation has the best detection performance (red line).

%% file: sections/7_discussion.tex
\subsection{Implementation costs}
\label{sec:diss}
\begin{table}[tpb]
 \centering
\scalebox{0.8}{
\setlength{\tabcolsep}{1mm}{
\begin{tabular}{c c|c|c|c|c}
\hline
&\textbf{Model}          & \textbf{FLOPs(G)} & \textbf{Params(M)} & \textbf{Time(s)} &\textbf{Overall}\\ \hline

 \multirow{4}{*}{\rotatebox{90}{\textbf{\makecell[c]{ATC}}}} & R2021Fixing70   & 38.8              & 254.44             & 1.21       & 11945     \\ \cline{2-6}
&G2021Improving70  & 38.8              & 254.44             & 1.21     & 11945      \\ \cline{2-6}
&G2020Uncovering70  & 38.8              & 254.44             & \underline{\textit{1.21}}    & 11945   \\   \cline{2-6}
&R2021Fixing106  & 60.57             & 396.23             & 1.24       & 29760    \\ \hhline{==|=|=|=|=}

\multirow{2}{*}{\rotatebox{90}{\textbf{\makecell[c]{\footnotesize{Det.}}}}}&LNG                     & \textbf{0.28}             & \textbf{7.95}               & 9.22  &  \underline{\textit{20.521}}       \\ \cline{2-6}
&\textbf{\beyond}                  & \underline{\textit{0.715}}             & \underline{\textit{19.67}}              & \textbf{1.12} &  \textbf{15.75}          \\ \hline
\end{tabular}}}
\caption{Comparison of time,computational and storage costs.}
\label{tb:flops}
\end{table}

\beyond uses an additional SSL model for AE detection, which inevitably increases the computational and storage cost. And the inference time (speed) is also considered in practice. 
Tab. \ref{tb:flops} presents the comparison for SOTA adversarial training defense and AE detection method, i.e. LNG. 
The detection models have a smaller model structure than those of ATCs, which can be reflected by the \textit{Params} and \textit{FLOPs}~\cite{flops} being much lower than those of ATC.
For \beyond, the projection head is a three-layer FC, leading to higher parameters and \textit{FLOPs} than LNG. 
However, \beyond only compares the relationship between neighbors without calculating the distance with the reference set, resulting in a faster inference speed than that of LNG. We show the 
FLOPs $\times$ Params $\times$ Time as the \textit{Overall} metric in Tab.~\ref{tb:flops}'s last column for overall comparison. If cost is a real concern in some scenarios, we can further reduce the cost with some strategy, e.g., reducing the neighbor number, without compromising performance significantly, as shown in Fig.~\ref{fig:k_pgd}.

%% file: sections/8_conclusion.tex
\section{Conclusion}
\label{sec:conclusion}
In this paper, we take the first step to detect AEs by identifying abnormal relations between AEs and their neighbors without prior knowledge of AEs. Samples have low label consistency, and representation similarity with their neighbors is detected as AE. Experiments with gray-box and white-box attacks show that \beyond outperforms the SOTA AE detectors in both detection ability and efficiency. Moreover, as a plug-and-play model, \beyond can be well combined with ATC to further improve robustness.

%% file: main.bbl
\begin{thebibliography}{57}
\providecommand{\natexlab}[1]{#1}

\bibitem[{Abusnaina et~al.(2021)Abusnaina, Wu, Arora, Wang, Wang, Yang, and
  Mohaisen}]{lng}
Abusnaina, A.; Wu, Y.; Arora, S.; Wang, Y.; Wang, F.; Yang, H.; and Mohaisen,
  D. 2021.
\newblock Adversarial example detection using latent neighborhood graph.
\newblock In \emph{Proceedings of the IEEE/CVF International Conference on
  Computer Vision}, 7687--7696.

\bibitem[{Andriushchenko et~al.(2020)Andriushchenko, Croce, Flammarion, and
  Hein}]{square}
Andriushchenko, M.; Croce, F.; Flammarion, N.; and Hein, M. 2020.
\newblock Square Attack: A Query-Efficient Black-Box Adversarial Attack via
  Random Search.
\newblock In \emph{Computer Vision – ECCV 2020: 16th European Conference,
  Glasgow, UK, August 23–28, 2020, Proceedings, Part XXIII}, 484–501.
  Berlin, Heidelberg: Springer-Verlag.
\newblock ISBN 978-3-030-58591-4.

\bibitem[{Athalye, Carlini, and Wagner(2018)}]{obfuscated}
Athalye, A.; Carlini, N.; and Wagner, D. 2018.
\newblock Obfuscated gradients give a false sense of security: Circumventing
  defenses to adversarial examples.
\newblock In \emph{International conference on machine learning}, 274--283.
  PMLR.

\bibitem[{Athalye et~al.(2018)Athalye, Engstrom, Ilyas, and Kwok}]{eot}
Athalye, A.; Engstrom, L.; Ilyas, A.; and Kwok, K. 2018.
\newblock Synthesizing robust adversarial examples.
\newblock In \emph{International conference on machine learning}, 284--293.
  PMLR.

\bibitem[{Bhatia and Davis(1995)}]{bhatia1995cauchy}
Bhatia, R.; and Davis, C. 1995.
\newblock A Cauchy-Schwarz inequality for operators with applications.
\newblock \emph{Linear algebra and its applications}, 223: 119--129.

\bibitem[{Bryniarski et~al.(2021)Bryniarski, Hingun, Pachuca, Wang, and
  Carlini}]{opgd}
Bryniarski, O.; Hingun, N.; Pachuca, P.; Wang, V.; and Carlini, N. 2021.
\newblock Evading adversarial example detection defenses with orthogonal
  projected gradient descent.
\newblock \emph{arXiv preprint arXiv:2106.15023}.

\bibitem[{Carlini and Wagner(2017)}]{cw}
Carlini, N.; and Wagner, D. 2017.
\newblock Towards evaluating the robustness of neural networks.
\newblock In \emph{2017 ieee symposium on security and privacy (sp)}, 39--57.
  Ieee.

\bibitem[{Chen and He(2021)}]{simsiam}
Chen, X.; and He, K. 2021.
\newblock Exploring simple siamese representation learning.
\newblock In \emph{Proceedings of the IEEE/CVF Conference on Computer Vision
  and Pattern Recognition}, 15750--15758.

\bibitem[{Cococcioni et~al.(2020)Cococcioni, Rossi, Ruffaldi, Saponara, and
  de~Dinechin}]{autodrive}
Cococcioni, M.; Rossi, F.; Ruffaldi, E.; Saponara, S.; and de~Dinechin, B.~D.
  2020.
\newblock Novel arithmetics in deep neural networks signal processing for
  autonomous driving: Challenges and opportunities.
\newblock \emph{IEEE Signal Processing Magazine}, 38(1): 97--110.

\bibitem[{Croce et~al.(2020)Croce, Andriushchenko, Sehwag, Debenedetti,
  Flammarion, Chiang, Mittal, and Hein}]{robustbench}
Croce, F.; Andriushchenko, M.; Sehwag, V.; Debenedetti, E.; Flammarion, N.;
  Chiang, M.; Mittal, P.; and Hein, M. 2020.
\newblock Robustbench: a standardized adversarial robustness benchmark.
\newblock \emph{arXiv preprint arXiv:2010.09670}.

\bibitem[{Croce et~al.(2022)Croce, Gowal, Brunner, Shelhamer, Hein, and
  Cemgil}]{eval_robust}
Croce, F.; Gowal, S.; Brunner, T.; Shelhamer, E.; Hein, M.; and Cemgil, T.
  2022.
\newblock Evaluating the Adversarial Robustness of Adaptive Test-time Defenses.
\newblock \emph{arXiv preprint arXiv:2202.13711}.

\bibitem[{Croce and Hein(2020)}]{autoattack}
Croce, F.; and Hein, M. 2020.
\newblock Reliable evaluation of adversarial robustness with an ensemble of
  diverse parameter-free attacks.
\newblock In \emph{International conference on machine learning}, 2206--2216.
  PMLR.

\bibitem[{da~Costa et~al.(2022)da~Costa, Fini, Nabi, Sebe, and Ricci}]{solo}
da~Costa, V. G.~T.; Fini, E.; Nabi, M.; Sebe, N.; and Ricci, E. 2022.
\newblock solo-learn: A Library of Self-supervised Methods for Visual
  Representation Learning.
\newblock \emph{J. Mach. Learn. Res.}, 23: 56--1.

\bibitem[{Demontis et~al.(2019)Demontis, Melis, Pintor, Jagielski, Biggio,
  Oprea, Nita-Rotaru, and Roli}]{demontis2019adversarial}
Demontis, A.; Melis, M.; Pintor, M.; Jagielski, M.; Biggio, B.; Oprea, A.;
  Nita-Rotaru, C.; and Roli, F. 2019.
\newblock Why do adversarial attacks transfer? explaining transferability of
  evasion and poisoning attacks.
\newblock In \emph{28th USENIX security symposium (USENIX security 19)},
  321--338.

\bibitem[{Dubey et~al.(2019)Dubey, van~der Maaten, Yalniz, Li, and
  Mahajan}]{knn}
Dubey, A.; van~der Maaten, L.; Yalniz, I.~Z.; Li, Y.; and Mahajan, D. 2019.
\newblock Defense Against Adversarial Images using Web-Scale Nearest-Neighbor
  Search.
\newblock \emph{computer vision and pattern recognition}.

\bibitem[{Elfwing, Uchibe, and Doya(2018)}]{sigmoid}
Elfwing, S.; Uchibe, E.; and Doya, K. 2018.
\newblock Sigmoid-weighted linear units for neural network function
  approximation in reinforcement learning.
\newblock \emph{Neural Networks}, 107: 3--11.

\bibitem[{Gowal et~al.(2020)Gowal, Qin, Uesato, Mann, and Kohli}]{uncovering}
Gowal, S.; Qin, C.; Uesato, J.; Mann, T.; and Kohli, P. 2020.
\newblock Uncovering the limits of adversarial training against norm-bounded
  adversarial examples.
\newblock \emph{arXiv preprint arXiv:2010.03593}.

\bibitem[{Gowal et~al.(2021)Gowal, Rebuffi, Wiles, Stimberg, Calian, and
  Mann}]{improving}
Gowal, S.; Rebuffi, S.-A.; Wiles, O.; Stimberg, F.; Calian, D.~A.; and Mann,
  T.~A. 2021.
\newblock Improving robustness using generated data.
\newblock \emph{Advances in Neural Information Processing Systems}, 34:
  4218--4233.

\bibitem[{He et~al.(2016)He, Zhang, Ren, and Sun}]{resnet}
He, K.; Zhang, X.; Ren, S.; and Sun, J. 2016.
\newblock Deep residual learning for image recognition.
\newblock In \emph{Proceedings of the IEEE conference on computer vision and
  pattern recognition}, 770--778.

\bibitem[{Hendrycks et~al.(2019{\natexlab{a}})Hendrycks, Mazeika, Kadavath, and
  Song}]{hendrycks2019using}
Hendrycks, D.; Mazeika, M.; Kadavath, S.; and Song, D. 2019{\natexlab{a}}.
\newblock Using self-supervised learning can improve model robustness and
  uncertainty.
\newblock \emph{Advances in neural information processing systems}, 32.

\bibitem[{Hendrycks et~al.(2019{\natexlab{b}})Hendrycks, Mazeika, Kadavath, and
  Song}]{ssl_robust}
Hendrycks, D.; Mazeika, M.; Kadavath, S.; and Song, D. 2019{\natexlab{b}}.
\newblock Using self-supervised learning can improve model robustness and
  uncertainty.
\newblock \emph{Advances in neural information processing systems}, 32.

\bibitem[{Ho and Nvasconcelos(2020)}]{ssl_at2}
Ho, C.-H.; and Nvasconcelos, N. 2020.
\newblock Contrastive learning with adversarial examples.
\newblock \emph{Advances in Neural Information Processing Systems}, 33:
  17081--17093.

\bibitem[{Hu et~al.(2019)Hu, Yu, Guo, Chao, and Weinberger}]{hu}
Hu, S.; Yu, T.; Guo, C.; Chao, W.-L.; and Weinberger, K.~Q. 2019.
\newblock A New Defense Against Adversarial Images: Turning a Weakness into a
  Strength.
\newblock \emph{neural information processing systems}.

\bibitem[{Jaiswal et~al.(2020)Jaiswal, Ramesh~Babu, Zaki~Zadeh, Banerjee, and
  Makedon}]{jaiswal2020survey}
Jaiswal, A.; Ramesh~Babu, A.; Zaki~Zadeh, M.; Banerjee, D.; and Makedon, F.
  2020.
\newblock A Survey on Contrastive Self-Supervised Learning.

\bibitem[{Jiang et~al.(2020)Jiang, Chen, Chen, and Wang}]{feauture_invar}
Jiang, Z.; Chen, T.; Chen, T.; and Wang, Z. 2020.
\newblock Robust pre-training by adversarial contrastive learning.
\newblock \emph{Advances in Neural Information Processing Systems}, 33:
  16199--16210.

\bibitem[{Kaissis et~al.(2022)Kaissis, Makowski, R{\"u}ckert, and
  Braren}]{disease}
Kaissis, G.; Makowski, M.; R{\"u}ckert, D.; and Braren, R. 2022.
\newblock Secure, privacy-preserving and federated machine learning in medical
  imaging.

\bibitem[{Kim, Tack, and Hwang(2020)}]{ssl_at1}
Kim, M.; Tack, J.; and Hwang, S.~J. 2020.
\newblock Adversarial self-supervised contrastive learning.
\newblock \emph{Advances in Neural Information Processing Systems}, 33:
  2983--2994.

\bibitem[{Krizhevsky, Hinton et~al.(2009)}]{cifar10}
Krizhevsky, A.; Hinton, G.; et~al. 2009.
\newblock Learning multiple layers of features from tiny images.

\bibitem[{Krizhevsky, Sutskever, and Hinton(2012)}]{imagenet}
Krizhevsky, A.; Sutskever, I.; and Hinton, G.~E. 2012.
\newblock Imagenet classification with deep convolutional neural networks.
\newblock \emph{Advances in neural information processing systems}, 25.

\bibitem[{Liu et~al.(2019)Liu, Zhang, Zhang, Hou, Liu, Zha, and Yu}]{spam}
Liu, J.; Zhang, W.; Zhang, Y.; Hou, D.; Liu, Y.; Zha, H.; and Yu, N. 2019.
\newblock Detection based defense against adversarial examples from the
  steganalysis point of view.
\newblock In \emph{Proceedings of the IEEE/CVF Conference on Computer Vision
  and Pattern Recognition}, 4825--4834.

\bibitem[{Liu et~al.(2021)Liu, Zhang, Hou, Mian, Wang, Zhang, and
  Tang}]{liu2021self}
Liu, X.; Zhang, F.; Hou, Z.; Mian, L.; Wang, Z.; Zhang, J.; and Tang, J. 2021.
\newblock Self-supervised learning: Generative or contrastive.
\newblock \emph{IEEE Transactions on Knowledge and Data Engineering}.

\bibitem[{Ma et~al.(2018)Ma, Li, Wang, Erfani, Wijewickrema, Schoenebeck, Song,
  Houle, and Bailey}]{lid}
Ma, X.; Li, B.; Wang, Y.; Erfani, S.~M.; Wijewickrema, S.; Schoenebeck, G.;
  Song, D.; Houle, M.~E.; and Bailey, J. 2018.
\newblock Characterizing adversarial subspaces using local intrinsic
  dimensionality.
\newblock \emph{arXiv preprint arXiv:1801.02613}.

\bibitem[{Madry et~al.(2017)Madry, Makelov, Schmidt, Tsipras, and Vladu}]{pgd}
Madry, A.; Makelov, A.; Schmidt, L.; Tsipras, D.; and Vladu, A. 2017.
\newblock Towards deep learning models resistant to adversarial attacks.
\newblock \emph{arXiv preprint arXiv:1706.06083}.

\bibitem[{Mao et~al.(2021)Mao, Chiquier, Wang, Yang, and Vondrick}]{purify2}
Mao, C.; Chiquier, M.; Wang, H.; Yang, J.; and Vondrick, C. 2021.
\newblock Adversarial attacks are reversible with natural supervision.
\newblock In \emph{Proceedings of the IEEE/CVF International Conference on
  Computer Vision}, 661--671.

\bibitem[{Mao et~al.(2020)Mao, Gupta, Nitin, Ray, Song, Yang, and
  Vondrick}]{multitask}
Mao, C.; Gupta, A.; Nitin, V.; Ray, B.; Song, S.; Yang, J.; and Vondrick, C.
  2020.
\newblock Multitask learning strengthens adversarial robustness.
\newblock In \emph{European Conference on Computer Vision}, 158--174. Springer.

\bibitem[{Meng and Chen(2017)}]{magnet}
Meng, D.; and Chen, H. 2017.
\newblock Magnet: a two-pronged defense against adversarial examples.
\newblock In \emph{Proceedings of the 2017 ACM SIGSAC conference on computer
  and communications security}, 135--147.

\bibitem[{Miko{\l}ajczyk and Grochowski(2018)}]{mikolajczyk2018data}
Miko{\l}ajczyk, A.; and Grochowski, M. 2018.
\newblock Data augmentation for improving deep learning in image classification
  problem.
\newblock In \emph{2018 international interdisciplinary PhD workshop (IIPhDW)},
  117--122. IEEE.

\bibitem[{Papernot and McDaniel(2018)}]{dknn}
Papernot, N.; and McDaniel, P. 2018.
\newblock Deep k-nearest neighbors: Towards confident, interpretable and robust
  deep learning.
\newblock \emph{arXiv preprint arXiv:1803.04765}.

\bibitem[{Papernot, McDaniel, and
  Goodfellow(2016)}]{papernot2016transferability}
Papernot, N.; McDaniel, P.; and Goodfellow, I. 2016.
\newblock Transferability in machine learning: from phenomena to black-box
  attacks using adversarial samples.
\newblock \emph{arXiv preprint arXiv:1605.07277}.

\bibitem[{Raff et~al.(2019)Raff, Sylvester, Forsyth, and
  McLean}]{raff2019barrage}
Raff, E.; Sylvester, J.; Forsyth, S.; and McLean, M. 2019.
\newblock Barrage of random transforms for adversarially robust defense.
\newblock In \emph{Proceedings of the IEEE/CVF Conference on Computer Vision
  and Pattern Recognition}, 6528--6537.

\bibitem[{Rebuffi et~al.(2021)Rebuffi, Gowal, Calian, Stimberg, Wiles, and
  Mann}]{fixing}
Rebuffi, S.-A.; Gowal, S.; Calian, D.~A.; Stimberg, F.; Wiles, O.; and Mann, T.
  2021.
\newblock Fixing data augmentation to improve adversarial robustness.
\newblock \emph{arXiv preprint arXiv:2103.01946}.

\bibitem[{Shan et~al.(2020)Shan, Wenger, Wang, Li, Zheng, and Zhao}]{trapdoor}
Shan, S.; Wenger, E.; Wang, B.; Li, B.; Zheng, H.; and Zhao, B.~Y. 2020.
\newblock Gotta Catch'Em All: Using Honeypots to Catch Adversarial Attacks on
  Neural Networks.
\newblock \emph{computer and communications security}.

\bibitem[{Shi, Holtz, and Mishne(2021)}]{purify1}
Shi, C.; Holtz, C.; and Mishne, G. 2021.
\newblock Online adversarial purification based on self-supervision.
\newblock \emph{arXiv preprint arXiv:2101.09387}.

\bibitem[{Sitawarin, Golan-Strieb, and Wagner(2022)}]{rt}
Sitawarin, C.; Golan-Strieb, Z.~J.; and Wagner, D. 2022.
\newblock Demystifying the Adversarial Robustness of Random Transformation
  Defenses.
\newblock In \emph{International Conference on Machine Learning}, 20232--20252.
  PMLR.

\bibitem[{Sperl et~al.(2020)Sperl, Kao, Chen, Lei, and B{\"o}ttinger}]{dla}
Sperl, P.; Kao, C.-Y.; Chen, P.; Lei, X.; and B{\"o}ttinger, K. 2020.
\newblock DLA: dense-layer-analysis for adversarial example detection.
\newblock In \emph{2020 IEEE European Symposium on Security and Privacy
  (EuroS\&P)}, 198--215. IEEE.

\bibitem[{Svoboda et~al.(2018)Svoboda, Masci, Monti, Bronstein, and
  Guibas}]{peernet}
Svoboda, J.; Masci, J.; Monti, F.; Bronstein, M.~M.; and Guibas, L. 2018.
\newblock Peernets: Exploiting peer wisdom against adversarial attacks.
\newblock \emph{arXiv preprint arXiv:1806.00088}.

\bibitem[{Szegedy et~al.(2013)Szegedy, Zaremba, Sutskever, Bruna, Erhan,
  Goodfellow, and Fergus}]{intriguing}
Szegedy, C.; Zaremba, W.; Sutskever, I.; Bruna, J.; Erhan, D.; Goodfellow, I.;
  and Fergus, R. 2013.
\newblock Intriguing properties of neural networks.
\newblock \emph{arXiv preprint arXiv:1312.6199}.

\bibitem[{Tian et~al.(2021)Tian, Zhou, Li, and Duan}]{sid}
Tian, J.; Zhou, J.; Li, Y.; and Duan, J. 2021.
\newblock Detecting adversarial examples from sensitivity inconsistency of
  spatial-transform domain.
\newblock In \emph{Proceedings of the AAAI Conference on Artificial
  Intelligence}, volume~35, 9877--9885.

\bibitem[{Tramer(2022)}]{detect_matter}
Tramer, F. 2022.
\newblock Detecting adversarial examples is (nearly) as hard as classifying
  them.
\newblock In \emph{International Conference on Machine Learning}, 21692--21702.
  PMLR.

\bibitem[{Tram{\`e}r et~al.(2017)Tram{\`e}r, Papernot, Goodfellow, Boneh, and
  McDaniel}]{visualization}
Tram{\`e}r, F.; Papernot, N.; Goodfellow, I.; Boneh, D.; and McDaniel, P. 2017.
\newblock The space of transferable adversarial examples.
\newblock \emph{arXiv preprint arXiv:1704.03453}.

\bibitem[{Van~der Maaten and Hinton(2008)}]{tsne}
Van~der Maaten, L.; and Hinton, G. 2008.
\newblock Visualizing data using t-SNE.
\newblock \emph{Journal of machine learning research}, 9(11).

\bibitem[{Wu et~al.(2020)Wu, Wang, Xia, Bailey, and Ma}]{sgm}
Wu, D.; Wang, Y.; Xia, S.-T.; Bailey, J.; and Ma, X. 2020.
\newblock Skip connections matter: On the transferability of adversarial
  examples generated with resnets.
\newblock \emph{arXiv preprint arXiv:2002.05990}.

\bibitem[{Xie et~al.(2020)Xie, Tan, Gong, Wang, Yuille, and Le}]{flops}
Xie, C.; Tan, M.; Gong, B.; Wang, J.; Yuille, A.~L.; and Le, Q.~V. 2020.
\newblock Adversarial examples improve image recognition.
\newblock In \emph{Proceedings of the IEEE/CVF Conference on Computer Vision
  and Pattern Recognition}, 819--828.

\bibitem[{Xu, Evans, and Qi(2017)}]{fs}
Xu, W.; Evans, D.; and Qi, Y. 2017.
\newblock Feature squeezing: Detecting adversarial examples in deep neural
  networks.
\newblock \emph{arXiv preprint arXiv:1704.01155}.

\bibitem[{Yang et~al.(2022)Yang, Gao, Li, Lai, and Xu}]{Yang2022WhatYS}
Yang, Y.; Gao, R.; Li, Y.; Lai, Q.; and Xu, Q. 2022.
\newblock What You See is Not What the Network Infers: Detecting Adversarial
  Examples Based on Semantic Contradiction.
\newblock In \emph{Network and Distributed System Security Symposium (NDSS)}.

\bibitem[{Zeng et~al.(2020)Zeng, Qiu, Memmi, and Qiu}]{zeng2020data}
Zeng, Y.; Qiu, H.; Memmi, G.; and Qiu, M. 2020.
\newblock A data augmentation-based defense method against adversarial attacks
  in neural networks.
\newblock In \emph{International Conference on Algorithms and Architectures for
  Parallel Processing}, 274--289. Springer.

\bibitem[{Zhang et~al.(2019)Zhang, Yu, Jiao, Xing, El~Ghaoui, and
  Jordan}]{trade}
Zhang, H.; Yu, Y.; Jiao, J.; Xing, E.; El~Ghaoui, L.; and Jordan, M. 2019.
\newblock Theoretically principled trade-off between robustness and accuracy.
\newblock In \emph{International conference on machine learning}, 7472--7482.
  PMLR.

\end{thebibliography}
